\documentclass{article}
\usepackage[left=1in,right=1in,top=1in,bottom=1in]{geometry}
\usepackage{natbib}
\usepackage{xfrac}
\usepackage[utf8]{inputenc} 
\usepackage[T1]{fontenc}    
\usepackage{hyperref}       
\usepackage{url}            
\usepackage{booktabs}       
\usepackage{amsfonts}       
\usepackage{nicefrac}       
\usepackage{microtype}      
\usepackage{xcolor}         
\usepackage{multirow}

\usepackage{todonotes}
\usepackage{amssymb}
\usepackage{amsmath}
\usepackage{accents} 
\usepackage{dsfont} 
\usepackage[inline]{enumitem} 
\usepackage{wrapfig}
 \usepackage[ruled, algo2e, noend]{algorithm2e}
\allowdisplaybreaks
\usepackage{algorithm}

\newtheorem{thmdef}{Definition}

\newtheorem{thmclm}{Claim}

\newtheorem{thmprop}{Proposition}

\newtheorem{thmappdef}{Definition}

\newtheorem{thmapplem}{Lemma}

\newtheorem{thmappcol}{Corollary}

\newtheorem{thmappprop}{Proposition}

\newenvironment{thmproof}[1][Proof]{\begin{trivlist}
\item[\hskip \labelsep {\textit{#1.}}]}{\end{trivlist}}

\def\cP{\mathcal{P}}

\def\E{\mathbb{E}}
\def\indic{\mathds{1}}

\def\cD{\mathcal D}

\def\cX{\mathcal X}
\def\cY{\mathcal Y}

\def\cF{\mathcal F}
\def\cZ{\mathcal Z}

\def\e{\epsilon}

\def\cR{\mathcal{R}}
\def \R{\mathbb{R}}

\DeclareMathOperator*{\argmax}{arg\,max}

\def\mmd{\mathrm{MMD}}
\newcommand{\dr}[1]{#1^{\bullet}}

\newcommand{\uwidehat}[1]{%
  \mathpalette\douwidehat{#1}%
}

\makeatletter
\newcommand{\douwidehat}[2]{%
  \sbox0{$\m@th#1\widehat{\hphantom{#2}}$}%
  \sbox2{$\m@th#1x$}
  \sbox4{$\m@th#1#2$}
  \dimen0=\ht0
  \advance\dimen0 -.8\ht2
  \dimen2=\dp4
  \rlap{%
    \raisebox{\dimexpr\dimen0-\dimen2}{%
      \scalebox{1}[-1]{\box0}%
    }%
  }%
  {#2}%
}
\makeatother
\title{Partial identification of kernel based two sample tests with mismeasured data}
\date{}

\author{%
  Ron Nafshi \\
  Computer Science and Engineering\\
  University of Michigan, Ann Arbor\\
  \and
  Maggie Makar\thanks{Corresponding author} \\
  Computer Science and Engineering \\
  University of Michigan, Ann Arbor \\
  \texttt{mmakar@umich.edu} \\
}

\begin{document}

\maketitle

\begin{abstract}
Nonparametric two-sample tests such as the Maximum Mean Discrepancy (MMD) are often used to detect differences between two distributions in machine learning applications. However, the majority of existing literature assumes that error-free samples from the two distributions of interest are available.We relax this assumption and study the estimation of the MMD under $\epsilon$-contamination, where a possibly non-random $\epsilon$ proportion of one distribution is erroneously grouped with the other. We show that under $\epsilon$-contamination, the typical estimate of the MMD is unreliable. Instead, we study partial identification of the MMD, and characterize sharp upper and lower bounds that contain the true, unknown MMD. We propose a method to estimate these bounds, and show that it gives estimates that converge to the sharpest possible bounds on the MMD as sample size increases, with a convergence rate that is faster than alternative approaches. Using three datasets, we empirically validate that our approach is superior to the alternatives: it gives tight bounds with a low false coverage rate. 
\end{abstract}

\section{Introduction}
Nonparametric two-sample tests are powerful tools for measuring the difference between two distributions. 
The Maximum Mean Discrepancy (MMD) \cite{gretton2012kernel} has emerged as a particularly useful nonparametric two-sample test in machine learning literature. 
It has been widely used in robust predictive and reinforcement learning \cite{kumar2019stabilizing, makar2022causally, li2017mmd, oneto2020exploiting, veitch2021counterfactual,goldstein2022learning}, fairness applications \cite{prost2019toward,madras2018learning, makar2022fairness,louizos2015variational} and distributionally robust optimization \cite{staib2019distributionally, kirschner2020distributionally} among others. 
Despite its importance and widespread use, the majority of existing work using the MMD assumes that observed samples are measured without error. 
As we show in this work, if this assumption does not hold, the typical MMD estimate is unreliable. 

Here, we study the estimation of the MMD where one of the samples observed is measured with error. 
Specifically, we consider the $\e$-contamination mechanism, where a possibly non-random $\e$ proportion of one of the two variables is erroneously grouped with the other variable.
This mismeasurement mechanism arises in many important applications. For example, $\e$-contamination arises when trying to identify if there are biomarkers for Myocardial Infarction (MI). In this setting, we can use the MMD to detect differences in genome sequences between healthy individuals and patients with myocardial MI.
Detecting differences between the two groups is complicated due to undiagnosed ``silent'' MI cases. These silent MI cases represent $\e$-contamination that occurs non-randomly: women's MI cases are more likely to go undiagnosed compared to men \cite{merz2011yentl}. 

In this paper, we show that the typical $\mmd$ estimates are unreliable when the data is collected under the $\e$-contamination mechanism. 
Instead, we resort to a partial identification approach, where we estimate upper and lower bounds on the $\mmd$. 
We characterize upper and lower bounds that are credible, meaning that they contain the true unknown $\mmd$, and sharp, meaning they cannot be made tighter without additional assumptions. Importantly, these bounds are identifiable using the observed contaminated data and an estimate of $\e$. We develop an estimation approach to compute the upper and lower bounds and analyze its behavior in finite samples. Our analysis shows that our approach gives estimates that converge to the sharpest possible upper and lower bounds as the sample size increases at a rate faster than the alternatives. 

\textbf{Our contributions are summarized as follows}: 
\begin{enumerate*}[label=\textbf{(\arabic*)}]
    \item We show that under $\e$-contamination the typical estimates of the $\mmd$ are unreliable,  
    \item We characterize sharp upper and lower bounds on the unknown $\mmd$ that are identifiable using only the observed contaminated data, and an estimate of $\e$,  
    \item We propose an estimation approach to compute the upper and lower bound and analyze its behavior in finite samples showing that its convergence to the true upper and lower bounds depends on the sample size and the degree of contamination (i.e., the value of $\e$), 
    \item We apply our approach to 3 datasets showing that it achieves a superior performance compared to alternative approaches. 
\end{enumerate*}

\paragraph{Related work.}

The majority of existing work on nonparametric two-sample testing focuses on establishing statistically and computationally efficient and consistent estimators of the difference between two distributions under the assumption that the observed samples are error-free \cite{gretton2012kernel,gretton2009fast,schrabefficient,domingo2023compress}. However, analysis of the two-sample testing problem in settings where the data is missing or noisy is limited. To our knowledge, the only existing work that tackles this challenge is in the context of survival analysis, where the measurement error model arises from the classical right-censoring of the data \cite{fernandez2021reproducing}. By contrast, we study a different measurement error mechanism and suggest methods for partial identification of the $\mmd$. 

Measurement error in the context of comparing two distributions arises frequently in fairness literature. For example, \citet{kallus2022assessing} study settings where we wish to audit predictive models, testing if they encode information about protected class membership. They consider a setting where we only have access to an imperfect proxy of the protected class membership and show that typical fairness metrics such as demographic parity and equalized odds are not identifiable. Similar to this work, they develop methods for partial identification of these metrics. A key difference between \cite{kallus2022assessing} and the work we present here is that the former focuses on comparing a single moment (the mean) of two distributions whereas our work allows a more rigorous comparison of infinitely many moments of two distributions.   
We also stress that while the methods presented here could be used in a fairness context, they are more widely applicable to any setting where we wish to compare two distributions.

\section{Preliminaries}
Our goal is to test if two samples $X = \{x_i\}_{i}^n \sim P_X(X)$, ${Y= \{y_i\}_{i}^n \sim P_Y(Y)}$ are drawn from different distributions, i.e., if $P_X(X) = P_Y(Y)$. To simplify notation, we assume that the two samples have the same size $=n$, but stress that our results hold when the two samples have different sizes. The challenge we wish to address is that instead of observing $X, Y$, we observe $\e$-contaminated $X'$ and $Y'$, where a possibly non-random $\e$ proportion of one of the two variables is incorrectly grouped with the other for $0 < \e <1$. 
Without loss of generality, we assume that an $\e$-proportion of $X$ is incorrectly grouped with $Y$.
Specifically, let $C^* = \{c^*_i\}_i^m$, with $m = \lfloor \e n \rfloor$ be the unobserved subset of $X$ that is grouped with $Y$. We can express the distributions over the observed samples in relation to the true distributions and the unknown contaminated samples as follows: 
\begin{align*}
    & P_{Y'}(Y') = (1 - \alpha)P_{Y}(Y) + \alpha P_{C^*}(C^*) \quad \text{and} \quad
    P_{X'}(X')  = (1 + \tilde{\alpha}) P_{X}(X)  -  \tilde{\alpha} P_{C^*}(C^*), &
\end{align*}
where $\alpha = \e/(1 + \e)$ and $\tilde{\alpha} = \e/(1 - \e)$.
We do not make any additional assumptions about $P_{C^*}(C^*)$. Importantly, we do not assume that the contamination is random, meaning \textit{we do not assume} that $P_{C^*}(C^*) = P_{X'}(X') = P_X(X)$.

We assume that the value of $\e$ is known \emph{a priori}, or can be empirically estimated from other data sources. We use $\E_{P_A}[A]$ to denote the expectation of $A$ according to the distribution $P_A(A)$, $A \cup B$ to denote the union of the set $A$ and $B$, and $A \setminus B$ to denote the difference between the two sets $A$ and $B$. We use $\#(A)$ to denote the cardinality of the set $A$. We use $\cX'$ and $\cY'$ to denote the topological spaces of $X'$ and $Y'$ respectively. 

We focus on kernel two-sample tests, specifically, the Maximum Mean Discrepancy, $\mmd$ \citep{gretton2012kernel}. 

\begin{thmdef}\label{def:mmd}
For $Z \sim P_{Z}$, $Z' \sim P_{Z'}$, $\cF$ such that $\cF: \cZ \rightarrow \R$, and $k: \cZ \times \cZ \rightarrow \R$ with $k$ being a positive definite kernel matrix, the Maximum Mean Discrepancy is defined as
\begin{align}\label{def:mmd}
    \mmd(\cF, P_{Z}, P_{Z'}) = \sup\nolimits_{f \in \cF} \big(\E_{P_{Z}} f(Z) - \E_{P_{Z'}} f(Z')\big), 
\end{align}
 and the witness function $f^*$ is defined as the function attaining the supremum in expression~\eqref{def:mmd}, with
$
    f^*(t) = \E_{P_{Z}}[k(Z, t)] - \E_{P_{Z'}} [k(Z', t)], 
$
up to a normalization constant.
\end{thmdef}

When $\cF$ is set to be a general reproducing kernel Hilbert space (RKHS), the $\mmd$ defines a metric on probability distributions, and is equal to zero if and only if $P_{Z} =P_{Z'}$. Throughout, we fix $\cF$ to be the RKHS with $\|f\|_\mathcal{\cF} \leq 1$ for all $f \in \cF$ and drop $\cF$ from the $\mmd$ arguments to simplify notation. We use $k(z,z')$ to denote the reproducing kernel of $\cF$, and assume that $0 \leq k(x', y') \leq \kappa$ for all $x', y' \in \cX', \cY'$. 

\citet{gretton2012kernel}, showed that when there is no measurement error, the following empirical estimate of the $\mmd$ is unbiased: 
\begin{align}\label{eq:typical_est}
    \mmd(X, Y) = \frac{1}{n(n-1)} \sum_{i, j \not= i} k(x_i, x_j) + \frac{1}{n(n-1)}  \sum_{i, j \not= i} k(y_i, y_j) - \frac{2}{n^2} \sum_{i, j} k(x_i, y_i).
\end{align}
In the $\e$-contamination setting, without additional strong assumptions, the $\mmd$ estimate is unreliable, meaning $\mmd(X', Y')$ might not converge to $ \mmd(P_{X'}, P_{Y'})$. 
So instead we study partial identifiability of $\mmd(P_{X}, P_{Y})$. Meaning, our goal is to estimate credible and informative lower and upper bounds on the unknown $\mmd(P_{X}, P_{Y})$.
For those bounds to be informative, they should be \emph{sharp}, meaning they cannot be made tighter without any additional assumptions.

\section{Theory} \label{sec:theory}
Our goal is to estimate upper and lower bounds that reflect our uncertainty in the $\mmd$ due to measurement error. 

To proceed with our analysis, it is helpful to parameterize the $\mmd$ as function of the contaminated samples $C$. With some abuse of notation, for an arbitrary distribution $P_C$, we have that: 
\begin{align} \label{eq:mmdc}
    \mmd(P_{C}, P_{X'}, P_{Y'}) = \sup_{f \in \cF} \left[(1 - \e) \E_{P_{X'}}f(X') - (1 + \e) \E_{P_{Y'}}f(Y') + 2 \e \E_{P_{C}}f(C) \right], 
\end{align}
with $\mmd(P_{X}, P_{Y}) = \mmd(P_{C^*}, P_{X'}, P_{Y'})$.
Our first result categorizes the sharpest possible bounds that can be attained without additional assumptions. 
\begin{thmclm}\label{clm:sharp}
Let $(\cY', \Omega)$ be a measurable space with $Y' \in \cY'$ and let $\cP$ be all the probability distributions on $(\cY', \Omega)$. Define $\cP(\alpha)$ to be all the possible probability distributions over the unknown $C^*$, i.e., $\cP(\alpha) = \{ (P_{Y'}(Y') - (1 - \alpha) \varphi)/\alpha : \varphi \in \cP \}$, then the following bounds are sharp:
\begin{align*}
\inf_{P_C \in \cP(\alpha)}\mmd(P_C,  P_{X'}, P_{Y'}) \leq \mmd(P_{C^*},  P_{X'}, P_{Y'}) \leq \sup_{P_C \in \cP(\alpha)}\mmd(P_C,  P_{X'}, P_{Y'}), 
\end{align*}
\end{thmclm}
\begin{thmproof}
The proof follows from the fact that without any additional assumptions, $C^*$ can take on any values in $\cY'$, and hence its corresponding distribution can be any distribution over subsets of $Y'$ with measure $=\alpha$. This means that the sharpest possible upper (lower) bound must be defined with respect to distributions over $P_C$ that maximize (minimize) the $\mmd$. 
\end{thmproof}

We use $P_{\overline{C}}$ to denote the distribution that maximizes the third term in claim~\ref{clm:sharp} and define $P_{\underline{C}}$ similarly. Claim~\ref{clm:sharp} gives us a recipe for constructing empirical bounds on the true $\mmd(P_{C^*}, P_{X'}, P_{Y'})$. To get an estimate of the upper bound, we need to identify the values of $C$ that render $X' \cup C$ and $Y' \setminus C$ most dissimilar. For a lower bound, we need to identify values of $C$ that render $X' \cup C$ and $Y' \setminus C$ most similar. Unless otherwise noted, we will focus on the analysis of the upper bound of the $\mmd$ since the arguments for the lower bound are nearly identical. 

To get an objective to optimize, we further expand the empirical version of equation~\ref{eq:mmdc} to isolate the terms that depend on $C$, which gives us the empirical objective to optimize. As we show in Lemma~\ref{lem:decompose}, in order to estimate $\mmd(P_{\overline{C}},  P_{X'}, P_{Y'})$, we first need to identify $\widehat{C}$: 
\begin{align}\label{eq:optim}
    \nonumber \widehat{C} & = \argmax_{C \in Y', \#(C)=m} \psi(C, X', Y')  =  \argmax_{C \in Y' } \frac{(1 - \e)}{n} \sum_i \sum_j k(x'_i, c_j)  \\
    & - \frac{(1 + \e)}{n} \sum_i \sum_j k(y'_i, c_j) 
    + \frac{\e}{n} \sum_i \sum_{j \not= i} k(c_i, c_j).  
\end{align}
Note that optimizing $\psi$ under a cardinality constraint in this manner is a variation of the knapsack problem, a classic combinatorial NP-hard optimization problem.
Instead, we analyze approximation strategies in two regimes: when $\e$ can take on any value in [0,1] and when $\e$ is sufficiently close to $0$. 
Our analysis relies on analyzing the stability of the estimation algorithms \cite{bousquet2002stability}. 

\paragraph{Approximation strategy for $\boldsymbol{\e \in [0, 1]}$.}
For any value of $\e$, we can directly maximize equation~\ref{eq:optim}. Noting that: 
$\max_{C} \psi(C \in Y', X', Y') \leq \max_{C} \psi(C \in \cY', X', Y')$, 
we can utilize, for example, iterative optimization algorithms to estimate an approximate $\widehat{C}$. Specifically, 
\begin{align}\label{eq:optim_io}
    \widehat{C}_{\circ} =  \argmax_{C \in \cY', \#(C)=m} \psi(C, X', Y').
\end{align}
While many iterative optimization algorithms can be used to optimize equation~\ref{eq:optim_io}, we follow \citet{jitkrittum2016interpretable} in focusing on Quasi-Newton methods such as the L-BFGS-B algorithm \cite{byrd1995limited}. For this reason we refer to this iterative optimization approach as the Quasi-Newton optimization \textbf{QNO} approach. We stress that our anlaysis holds for any valid optimizatio approach.  

In proposition~\ref{prop:opt_main}, we study how fast the estimate based on $\widehat{C}_{\circ}$ converges to the true upper bound. 
\begin{thmprop}\label{prop:opt_main}
    For $\mmd(P_{\overline{C}}, P_{X'}, P_{Y'})$ as defined in claim~\ref{clm:sharp}, $\widehat{C}_{\circ}$ as defined in equation~\ref{eq:optim_io}, with $\#(\widehat{C}_{\circ}) = m$, $0 \leq k(x', y') \leq \kappa$ for all $x', y' \in \cX', \cY'$, we have that: 
\begin{align*}
    P_{X', Y'} \bigg\{ | \mmd(P_{\overline{C}}, P_{X'}, P_{Y'}) - \mmd(\widehat{C}_{\circ}, X', Y') | > b_0 + \varepsilon \bigg\} \leq 2\exp\bigg(\frac{ - \varepsilon^2 n}{b_1}\bigg), 
\end{align*}
for $b_0 = 4\sqrt{\kappa} (n^{-1/2} + \e m)$ and $b_1 = 2 \kappa((1 - \e)(1 - \e + \e m)^2  + (1 + \e)  (1 + \e + \e m)^2)$
\end{thmprop}
The proof for proposition~\ref{prop:opt_main} and all other statements are presented in the Appendix. The proposition shows that the rate of convergence of the empirical $\mmd$ defined with respect to $\widehat{C}_{\circ}$ to the sharp upper bound depends on the sample size, the value of $\e$ and the size of the contaminated set $m$. As $\e$ decreases, the estimated $\mmd(\widehat{C}_{\circ}, X', Y')$ converges faster to its population counterpart $\mmd(P_{\overline{C}}, P_{X'}, P_{Y'})$. At $\e =0$, we recover the convergence rate of the uncontaminated $\mmd$ (\citet{gretton2012kernel}, theorem 7). 
As expected, as the sample size increases, the estimate gets closer to its population counterpart. However, the $\e m$ term in the denominator of the exponent means that the rate of convergence depends unfavorably on the size of the contaminated sample. The next section addresses this issue.

\paragraph{Approximation strategy for a sufficiently small $\boldsymbol{\e}$.}
This approach relies on the fact that for a fixed $n$, and as $\e \rightarrow 0$ the third term in equation~\ref{eq:optim} vanishes. 
Specifically for $\e \approx 0$: 
\begin{align} \label{eq:approx_noe}
    \psi(C, X', Y')  \approx \frac{(1 - \e)}{n} \sum_i \sum_j k(x'_i, c_j) 
     -   \frac{(1 + \e)}{n} \sum_i \sum_j k(y'_i, c_j) 
     = \frac{1}{m} \sum_i \hat{f}'(c_i).
\end{align}
where $\hat{f}'$ is a weighted version of the empirical estimate of the witness function definted with respect to the observed contaminated samples. 
This means that for $\e$ close to 0, maximizing $\psi$ is equivalent to computing the value of the witness function for every sample in $Y'$, and then taking the subset with the highest values to be the estimate of $\widehat{C}$.
Consider the following estimate of $\widehat{C}$:
\begin{align}\label{eq:sd_optim}
    \widehat{C}_{\hat{\gamma}} = \{ y': \hat{f}'(y') \geq \hat{\gamma}\} 
    \text{ with } 
    \hat{\gamma} = q(\hat{f}'(Y'), 1-\alpha), 
\end{align}

where $q(\hat{f}'(Y'), 1-\alpha)$ is defined as the $1-\alpha$ quantile of $\hat{f}'(Y')$. That is,
$q(\hat{f}'(Y'), 1-\alpha) = \inf\{ \hat{f}'(y') \in \hat{f}'(Y'): (1-\alpha) < \text{CDF}(\hat{f}'(y'))\}$.
Equation~\ref{eq:sd_optim} describes taking the $y'$ samples with weighted witness function values in the top $1 - \alpha$ quantile as the candidates for contaminated samples. Next, we show that $\widehat{C}_{\hat{\gamma}}$ is a valid estimate of $\overline{C}$. 

\begin{thmprop}\label{prop:sd_asymp}
    Let $C_{{\gamma}}$ be the solution to equation~\ref{eq:sd_optim} as $n \rightarrow \infty$. For a sufficiently small $\e$, we have that $P_{C_{\gamma}}  = P_{\overline{C}}$, where $P_{\overline{C}}$ is defined as the distribution that maximizes the third term in claim~\ref{clm:sharp}.
\end{thmprop}

While the full proof is stated in the appendix, we find it helpful to highlight the key insight behind proposition~\ref{prop:sd_asymp}. 
The key insight here is that the distribution over $C_{\gamma}$ \textit{stochastically dominates} any other distribution over $Y'$ with respect to the transformation $f'(Y')$. Meaning, there exists no other distribution over a subset of $Y'$ with measure $\alpha$ that can give a larger $\E_C[f'(C)]$ than $\E_{C_\gamma}[f'(C_\gamma)]$. 
We note in passing that this construction extends the classical seminal work by \citet{horowitz1995identification} on estimation of population means using contaminated data to the nonparametric hypothesis test setting. We refer to this approach as the stochastic dominance (\textbf{SD}) approach. 

It remains to show that the estimate of the $\mmd$ defined with respect to $\widehat{C}_{\hat{\gamma}}$ as estimated using a \textit{finite sample} converges to the true upper bound. We do that in the following proposition. 
\begin{thmprop} \label{prop:sd_main}
    For $\mmd(P_{\overline{C}}, P_{X'}, P_{Y'})$ as defined in claim~\ref{clm:sharp}, $\widehat{C}_{\hat{\gamma}}$ as defined in equation~\ref{eq:sd_optim} and $\kappa$ such that $0 \leq k(x, y) \leq \kappa$ for all $x, y \in \cX$. Then as for a sufficiently small $\e$: 
    \begin{align*}
        P_{X', Y'} \bigg\{ | \mmd(P_{\overline{C}}, P_{X'}, P_{Y'}) - \mmd(\widehat{C}_{\hat{\gamma}}, X', Y') | > b_0 + \varepsilon \bigg\} \leq 2\exp\bigg(\frac{ - \varepsilon^2 n}{b_1}\bigg)
    \end{align*}
for $b_0 = 4( \kappa /n)^{\sfrac{1}{2}}(1 + \e)$ and $b_1 = 2 \kappa \big((1 -\e)^3 + (1+\e) (1 + 3 \e)^2\big)$
\end{thmprop}
The proposition shows that unlike QNO, SD avoids the unfavorable dependence on $m$ leading to faster convergence. Similar to proposition~\ref{prop:opt_main}, at $\e =0$, we recover the convergence rate of the uncontaminated $\mmd$.

The key advantage of SD over QNO is that it reduces the problem of estimating $\widehat{C}$ to estimating the quantile of the univariate distribution, $P_{f'(Y')}$, which is a single scalar. 
By contrast, the iterative optimization-based approach needs to identify an $m \times d$ matrix, with $d$ being the dimension of the data. While helpful, the SD approach is limited by the fact that it is a valid approximation only for $\e$ sufficiently close to 0. 
In the next section, we design an approach that extends the SD approach making it valid for any value of $\e$

\section{Approach} \label{sec:approach}
In this section, we describe our main approach to estimating tight and credible upper and lower bounds on the $\mmd$. 
Unless otherwise noted, we describe the estimation procedure for constructing the upper bound since the lower bound is nearly identical. 
Our strategy hinges on identifying $\widehat{C}$, an $m$-sized subset of $Y'$ which, when removed from $Y'$ and added to $X'$, would render $Y'$ most dissimilar to $X'$, giving us a valid estimate of the the upper bound on the unknown $\mmd(C^*, X', Y')$.
Estimating $\widehat{C}$ allows us to estimate $\mmd(\widehat{C}, X', Y')$ in a straightforward manner: we can simply substitute $\widehat{C}$ for $C$ in the empirical version of equation~\ref{eq:mmdc}.

Our main approach builds upon the SD approach studied in section~\ref{sec:theory} by addressing its main limitation: that it gives a valid estimate of $\widehat{C}_{\hat{\gamma}}$ only for $\e$ sufficiently close to 0. Our approach overcomes this limitation by dividing the task of estimating $\widehat{C}_{\hat{\gamma}}$ into multiple, easier tasks each with an effective $\e^{(s)}$ that is smaller than the true $\e$. Specifically, we divide the estimation process into $S$ steps, in each step we estimate $\widehat{C}^{(s)}_{\hat{\gamma}^{(s)}}$, for $\e^{(s)} = \e/S$. 
Dividing the estimation into $S$ steps, with each step having $\e/S$-contamination means that each step of the estimation process will have an effective $\e$ that is close enough to 0 making equation~\ref{eq:sd_optim} a valid approximation, and overcoming the main limitation of SD. 
In the step $s$ of our algorithm, we calculate $\widehat{C}^{(s)}_{\hat{\gamma}^{(s)}} = \{ y' \in \widehat{Y}^{(s)}: \hat{f}^{(s)}(\widehat{Y}^{(s)}) \geq \hat{\gamma}^{(s)}\}$, for ${\hat{\gamma}^{(s)} = q(\hat{f}^{(s)}(\widehat{Y}^{(s)}), 1-\alpha^{(s)})}$ for $\alpha^{(s)} = \e^{(s)}/(1+ \e^{(s)})$, where 
\begin{align}\label{eq:step_witness}
    \hat{f}^{(s)}(\widehat{Y}^{(s)}) = \Big(1 - \frac{\e}{S}\Big) \frac{1}{n} \sum_i \sum_j k(\hat{x}^{(s)}_i, \hat{y}^{(s)}_j) 
-   \Big(1 + \frac{\e}{S}\Big)\frac{1}{n} \sum_i \sum_j k(\hat{y}^{(s)}_i, \hat{y}^{(s)}_j), 
\end{align}
with $\widehat{Y}^{(s)} = Y' \setminus \{\widehat{C}^{(1)}_{\hat{\gamma}^{(1)}}, \widehat{C}^{(2)}_{\hat{\gamma}^{(2)}}, \hdots \widehat{C}^{(s-1)}_{\hat{\gamma}^{(s-1)}}\}$, and $\widehat{X}^{(s-1)} = X' \cup  \{\widehat{C}^{(1)}_{\hat{\gamma}^{(1)}}, \widehat{C}^{(2)}_{\hat{\gamma}^{(2)}}, \hdots \widehat{C}^{(s-1)}_{\hat{\gamma}^{(s-1)}}\}$.

We refer to our Stepwise Stochastic Dominance based approach as \textbf{S-SD}. We summarize our procedure for estimating the upper and lower bounds in algorithms~\ref{alg:upper} and~\ref{alg:lower} respectively. We use $\uwidehat{C}$ to denote the counterpart of $\widehat{C}$ defined with respect to the lower bound. 
\begin{figure}
\begin{minipage}[t]{0.46\textwidth}
  \vspace{-1em}  
\begin{algorithm}[H]
 \KwIn{$X', Y', \e, S$}
 $\widehat{C} := \{\}$, $\alpha^{(s)} = \e/(\e + S)$\;\\
 \For{$s = 1 \hdots S$}{
   \ \  $X^{(s)} = X' \cup \widehat{C}$, $Y^{(s)} = Y' \setminus \widehat{C}$\; \\
   Compute $\hat{f}^{(s)}(Y^{(s)})$ as per equation~\ref{eq:step_witness}\\
   $\hat{\gamma}_{(1 - \e)} = q(\hat{f}^{(s)}(Y^{(s)}), 1 - \alpha^{(s)})$\\   
   $\widehat{C}^s = \{ y^{(s)}: \hat{f}^{(s)}(y^{(s)}) \geq \hat{\gamma}_{(1 - \e)}\}$\\
  $\widehat{C} := \widehat{C} \cup \widehat{C}^s$
 }
 \KwRet{$\mmd(\widehat{C}, X', Y')$}
 \caption{Our approach (S-SD) for estimating upper bounds}\label{alg:upper}
\end{algorithm}
\end{minipage}%
\hfill
\begin{minipage}[t]{0.46\textwidth}
\vspace{-1em} 
\begin{algorithm}[H]
 \KwIn{$X', Y', \e, S$}
 $\uwidehat{C} := \{\}$, $\alpha^{(s)} = \e/(\e + S)$\;\\
 \For{$s = 1 \hdots S$}{
   \ \  $X^{(s)} = X' \cup \uwidehat{C}$, $Y^{(s)} = Y' \setminus \uwidehat{C}$\; \\
   Compute $\hat{f}^{(s)}(Y^{(s)})$ as per equation~\ref{eq:step_witness}\\
   $\hat{\gamma}_{\e} = q(\hat{f}^{(s)}(Y^{(s)}), \alpha^{(s)})$\\ 
   $\uwidehat{C}^s = \{ y^{(s)}: \hat{f}^{(s)}(y^{(s)}) \leq \hat{\gamma}_{\e}\}$\\
  $\uwidehat{C} := \uwidehat{C} \cup \uwidehat{C}^s$
 }
 \KwRet{$\mmd(\uwidehat{C}, X', Y')$}
 \caption{Our approach (S-SD) for estimating lower bounds}\label{alg:lower}
\end{algorithm}

\end{minipage}
\end{figure}
We note that $S$ is a user-specified parameter that takes on values between 0 and $m$. 
In section~\ref{sec:step_sens} we give practical guidance on how to set $S$. 


\section{Experiments} \label{sec:exp}

In this section, we analyze the credibility and tightness of our approach and baselines using the False Coverage Rate (FCR) and Mean Interval Width (MIW) respectively. For $L$ draws of $X', Y'$ each of size $(1 - \e)n$ and $(1+\e)n$ respectively, the FCR and the MIW are defined as follows: 
\begin{align*}
    \text{FCR} & = \frac{1}{L} \sum_i \indic\{\mmd(\uwidehat{C}, X'_i, Y'_i) \leq \mmd(C^*, X'_i, Y'_i) \leq \mmd(\widehat{C}, X'_i, Y'_i)\} \\
    \text{MIW} & = \frac{1}{L} \sum_i | \mmd(\uwidehat{C}, X', Y') - \mmd(\widehat{C}, X', Y') | 
\end{align*}
We study the performance of our approach and baselines in three different datasets. Our aim is to study the effect of (1) varying data dimensions, (2) varying sample sizes, and (3) varying values of $\e$ on the performance of our approach as well as baselines.
In addition, we examine the sensitivity of our approach to varying the number of steps $S$.

\textbf{Ablations}. We study the following ablations of our approach:
\begin{enumerate*}[label=\textbf{(\arabic*)}]
    \item \textbf{SD}: For $S = 1$, S-SD becomes the same as SD. The performance of SD compared to S-SD highlights the importance of splitting the estimation procedure into $S$ steps.
    \item Stepwise-QNO (\textbf{S-QNO}): Follows the same steps outlined in algorithm~\ref{alg:upper}, however, instead of estimating $\widehat{C}^{(s)}_{\hat{\gamma}}$ and $\uwidehat{C}^{(s)}_{\hat{\gamma}}$ as a subroutine, it estimates $\widehat{C}^{(s)}_{\circ}$ and $\uwidehat{C}^{(s)}_{\circ}$ following equation~\ref{eq:optim} using the L-BFGS-B optimization algorithm. In each step $s$, this approach gives an estimate for an $m/S$ subset of candidate contaminated samples. This ablation study highlights the importance of using the SD approach as a subroutine. 
    \item \textbf{QNO}: Similar to S-QNO with $S=1$. 
\end{enumerate*}

\textbf{Baselines.} In addition to our main approach and the ablations, we investigate the following baselines: 
\begin{enumerate*}[label=\textbf{(\arabic*)}]
    \item Submodular optimization (\textbf{SM}): based on the approach suggested in \cite{kim2016examples}. It estimates $\widehat{C}$ by converting equation~\ref{eq:optim} into a submodular function by adding a submodular regularizer. Specifically, it greedily selects samples which maximise the function, $\max_{m} \hat{f}'(C) + \log \det k(C,C)$, where $\hat{f}'(C)$ is the witness function defined with respect to $X'$ and $Y'$, and $\log \det k(C,C)$ is the log-determinant regularizer.
    \item \textbf{Bootstrap}: a simple bootstrapping approach, which constructs bounds by resampling both observed groups with replacement and computing the $\mmd$ multiple times. The upper and lower bounds are then defined as the $(1-\alpha)$-th and $\alpha$ quantiles respectively over the distribution of resampled $\mmd$ values.
    The bootstrap estimates are centered around the typical $\mmd$ estimate (equation~\ref{eq:typical_est}), and hence they show how it behaves under $\e$-contamination \footnote{In the appendix, we explicitly show how the typical estimate of the $\mmd$ behaves with varying $\e$}.
\end{enumerate*}

For our approach, baselines and ablations, we fix the kernel to be the radial basis kernel (RBF) and use the median heuristic on the contaminated samples to determine bandwidth. Unless otherwise noted, we set the number of steps $S$ for S-SD and S-QNO to be $S=\min(m, 10)$; we take this minimum for when the total number of contaminated samples is less than the total number of steps. We examine the performance of different values of $S$ in section \ref{sec:step_sens}.

\textbf{Setup.} \label{Setup} Since the true value of the contaminated samples $C^*$ is unobserved in real datasets, we resort to semi-simulated data where $X, Y$ represent real data, but the contaminated samples are simulated. 
We examine the performance of our approach, ablations and baselines in two settings. First, is the nonrandom contamination setting. In this setting, we pick the data points that maximize the difference between the two distributions to be the true contaminated samples.
Specifically, we simulate contamination by randomly sampling $C^*$, a set of size $m$ from the $\min{\left(2m,n\right)}$ samples in $X$ with the largest witness function values, where the witness function here is defined with respect to the uncontaminated $X, Y$. 
We then create the observed samples $X' = X \setminus C^*$ and $Y' = Y \cup C^*$.  
Second, is the random contamination setting, where $C^*$ is sampled at random from $X$. 
Since the nonrandom contamination setting is more challenging, we present the results from that setting in the main text. Results from the random contamination setting are presented in the appendix.  
We define $N = \#(X) + \#(Y)$, the total number of samples, and consider 3 tasks corresponding to 3 datasets: 
\begin{enumerate}[wide, labelwidth=1pt, labelindent=0pt]
    \item \textbf{FOREST}: A publicly available dataset from the UCI KDD ML archive containing measurements of 54 cartographic variables such as elevation, slope, distance to water, and presence of certain sediment types \cite{forest}. We consider the task of performing a hypothesis test of habitat similarity for an ecological survey by estimating the $\mmd$ between the two forest types Lodgepole Pine and Spruce-Fir. We simulate $\e$ contamination by flipping an $\e$ proportion of Lodgepole Pine $(n=283,301)$ labels to Spruce-Fir $(n=211,840)$.

    \item \textbf{MIMIC}: A publicly available chest radiographs and corresponding clinical data with over 377,000 chest X-ray images and radiology reports \cite{johnson_pollard_mark_berkowitz_horng_2019,johnson_pollard_berkowitz_greenbaum_lungren_deng_mark_horng_2019,PhysioNet}. Here, we consider the task of testing if pneumonia predictions from a deep learning model trained on frontal chest x-rays depend on a sensitive attribute, such as the race of the patient. In this setting, the sensitive attribute is measured with $\e$-contamination. We use $60\%$ of the data for training the model, $20\%$ for validation, and the remaining $20\%$ for $\mmd$ testing. We use the training and validation data to fine tune a Densenet-121 \cite{Densenet} that was pretrained on Imagenet \cite{deng2009imagenet}. After training the model, we obtain the 2-dimensional logit predictions of the $20\%$ of the data held out for $\mmd$ testing, and simulate $\e$-contamination by changing an $\e$ proportion of Black $(n=3897)$ patients to White $(n=11293)$.
    
    \item \textbf{BIO}: Unlike the $2$-dimensional MIMIC data and $54$-dimensional FOREST data, in the third task we examine a more extreme case of high dimensional data with few samples. We use publicly available leukemia gene expression dataset (BIO) \cite{gloub}, which has 7128 measurements of gene expressions from DNA microarrays for 72 samples. The 72 samples are divided into binary groups of leukemia cancer cell types, acute lymphoblastic leukemia (ALL) and acute myeloid leukemia (AML), and we conduct the $\e$ contamination by flipping $\e$ of the ALL $(n=47)$ to AML $(n=25)$.
\end{enumerate}

\subsection{Performance under different data dimensions} \label{ex:samp_size}

\begin{table}[] 
\resizebox{\textwidth}{!}{%
\begin{tabular}{l|ll|ll|ll|}
\toprule
          & \multicolumn{2}{c|}{MIMIC ($N = 100, d = 2)$} & \multicolumn{2}{c|}{FOREST $(N = 100, d = 54)$} & \multicolumn{2}{c|}{BIO $(N = 72, d = 7128)$} \\
\cmidrule(r){2-3} \cmidrule(r){4-5} \cmidrule(r){6-7}
Approach  & FCR    & MIW   & FCR    & MIW & FCR    & MIW    \\
\cmidrule(r){2-3} \cmidrule(r){4-5} \cmidrule(r){6-7}
S-SD (Ours)     & $\mathbf{0.0 \pm (0.0)}$ & $\mathbf{0.137 \pm (0.008)}$      & $\mathbf{0.0 \pm (0.0)}$ & $\mathbf{0.088 \pm (0.003)}$     & $\mathbf{0.1 \pm (0.03)}$ & $\mathbf{0.075 \pm (0.001)}$ \\
S-QNO     & $0.08 \pm (0.067)$ & $0.119 \pm (0.006)$   & $0.02 \pm (0.02)$ & $0.084 \pm (0.004)$     & $1.0 \pm (0.0)$ & $0.059 \pm (0.001)$ \\
QNO              & $0.58 \pm (0.069)$ & $0.13 \pm (0.006)$    & $0.62 \pm (0.069)$ & $0.033 \pm (0.006)$     & $1.0 \pm (0.0)$ & $0.037 \pm (0.001)$ \\
SD               & $0.64 \pm (0.068)$ & $0.082 \pm (0.01)$   & $0.9 \pm (0.042)$ & $0.027 \pm (0.005)$      & $0.13 \pm (0.034)$ & $0.069 \pm (0.001)$ \\
SM               & $0.66 \pm (0.067)$ & $0.08 \pm (0.01)$    & $0.9 \pm (0.042)$ & $0.026 \pm (0.004)$      & $0.82 \pm (0.038)$ & $0.037 \pm (0.001)$ \\
Bootstrap        & $0.94 \pm (0.034)$ & $0.048 \pm (0.002)$  & $0.4 \pm (0.069)$ & $0.034 \pm (0.001)$      & $0.25 \pm (0.043)$ & $0.036 \pm (0.001)$ \\
\bottomrule
\end{tabular}
}
\caption{MIW and FCR for all datasets at $\e = 0.2$. Numbers in bold correspond to lowest FCR with smallest MIW.
Standard errors (in parentheses) computed by averaging over 100 trials. 
Results show that our approach performs better than all other approaches when the sample size is small and the dimension is large. In easier settings, our performs comparably to S-QNO.
}\label{nonrandom_main_table}
\end{table}

In this section, we examine the effect of varying dimension. To do so, we compute the FCR and MIW of bounds estimated on MIMIC $(N = 100, d = 2)$, FOREST $(N = 100, d = 54)$, and BIO $(N = 72, d = 7128)$ in table \ref{nonrandom_main_table}. We focus on the small sample regime as it is much more challenging. To get estimates for the standard error (SE) around the MIW and FCR, we repeat the experiment 100 times on 100 samples picked without replacement for MIMIC and FOREST. For BIO, we create 100 bootstrap samples. We fix $\e = 0.2$, simulate contamination in 100 random samples, and calculate the upper and lower bounds for each approach. 

The results in table \ref{nonrandom_main_table} show that in all settings our proposed approach gives the tightest (smallest MIW) and most credible (lowest FCR) estimates, while SD, QNO and S-QNO return bounds with a higher FCR. 
In settings where the dimensions are small, S-QNO performs significantly better than QNO. However, both perform poorly when the dimension, $d$ is large.
Such a finding makes sense: the stepwise algorithm reduces the dependence on the sample size, however the performance of both QNO and S-QNO appears to have some irreducible dependence on the dimension. 
This is not surprising, in BIO, for example, S-QNO is solving an optimization problem over an $m/S \times 7128$ parameter space, whereas S-SD is required to estimate the $(1- \alpha)/S$ quantile of a univariate distribution (that is the distribution over the values of the witness function). 
In this setting where $\e=0.2$, equation~\ref{eq:sd_optim} is a poor approximation of equation~\ref{eq:optim}, which explains the poor performance of SD. 
At $\e=0.2$ the typical estimate of the MMD (equation~\ref{eq:typical_est}) is unreliable. Being centered around the typical estimate, Bootstrap is expected to give unreliable bounds. 
SM also performs poorly since it is designed to find \textit{few} samples that explain the difference between the two corrupted distributions. 

Overall, S-SD remains robust even in high dimensions, while other approaches do not. In the appendix, we repeat this experiment with $N=2000$ for both MIMIC and FOREST. The results are largely consistent with the findings presented here. However, as $N$ increases, the estimates for S-QNO in small dimensions become more comparable to S-SD. 

For brevity, we present results on the FOREST dataset in the main text but include the similar analyses on MIMIC and BIO in the appendix.

\subsection{Performance under different sample sizes}

\begin{figure}[] 
\centering
  \includegraphics[width=\textwidth]{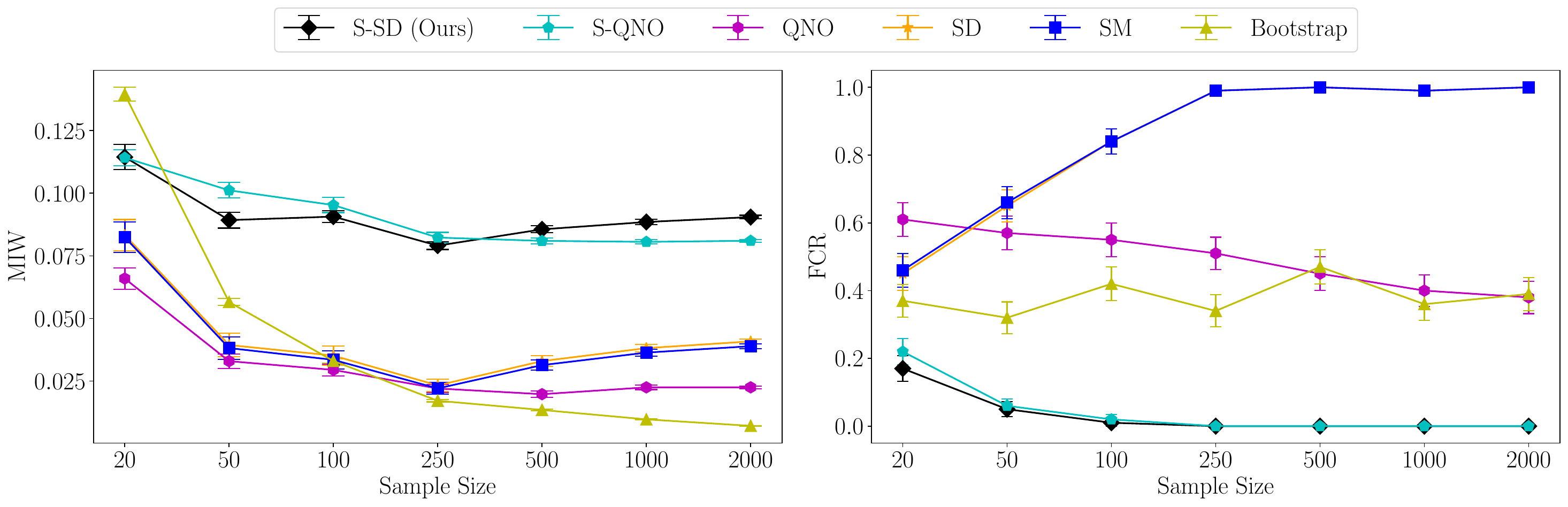}
  \vspace{-1.5em}
  \caption{Results on FOREST fixing $\e = 0.2$ and increasing sample size from $N = 20$ to $N = 2000$. Bars indicate the SE of the FCR and MIW across all trials. As sample size increases, MIW decreases for all methods, with S-SD providing intervals with the lowest FCR for all sample sizes.}\label{IW_FCR}
\end{figure}

Here, we focus on the effect of increasing sample size. Fixing $\e = 0.2$, we vary the sample size from $N= 20$ to $N = 2000$ by sampling from the FOREST dataset. For each sample size, we sample 100 times and compute the mean FCR and MIW and their corresponding standard errors. 
We plot the results for the MIW in figure~\ref{IW_FCR} (left) and the FCR in figure~\ref{IW_FCR} (right). 
The results show that the FCR for our approach, S-QNO and QNO decreases as the sample size increases revealing that these estimates are consistent. However, our approach gives the lowest FCR even in very small samples. In larger samples, S-QNO performs comparably to our approach. SD, SM and the bootstrap method all return overly conservative estimates that do not contain the true $\mmd$.

\subsection{Performance under different values of $\e$} \label{eps_increasing}

\begin{figure}
    \centering
    \includegraphics[width=\textwidth]{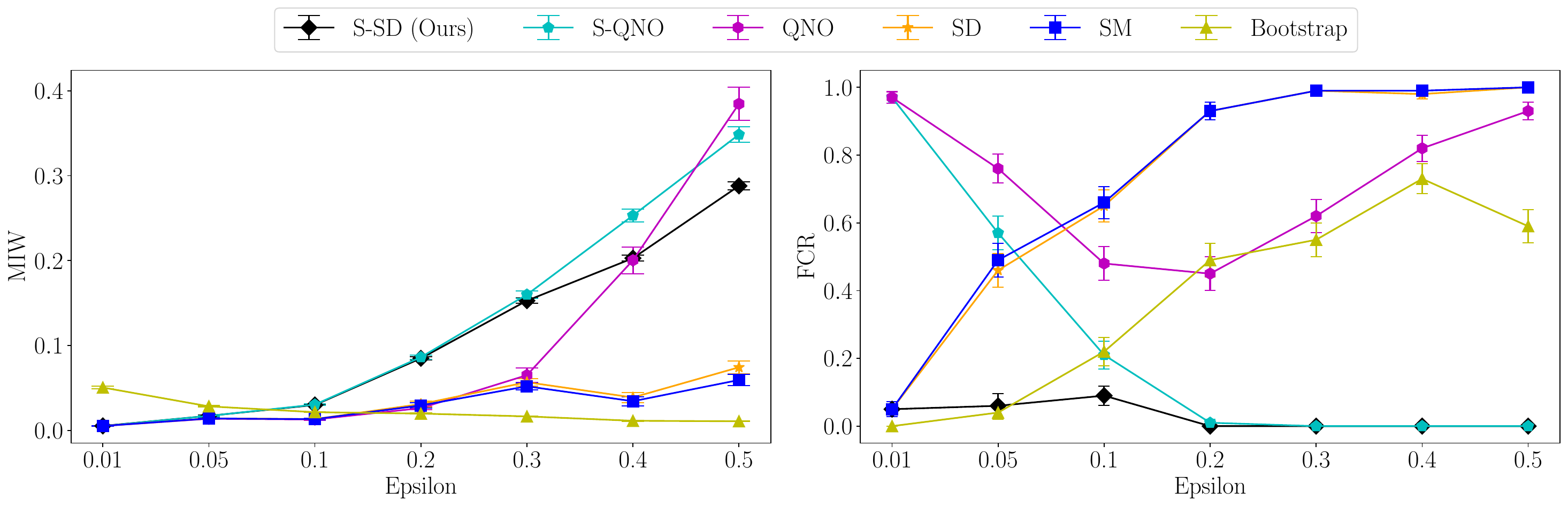}
    \vspace{-1.5em}
    \caption{The MIW and FCR for each approach is shown as the intensity of $\e$-contamination varies from $\e = 0.01$ to $\e = 0.5$ in FOREST $(N=100, d=54)$. Bars indicate the SE of the FCR and MIW across all trials. As $\e$ increases, S-SD reports tight and credible intervals for all values of $\e$.}
    \label{fig:FOREST_eps}
\end{figure}

Here, we investigate the effect of increasing contamination from $\e = 0.01$ to $\e = 0.9$.
Similar to section \ref{ex:samp_size}, we focus on the small sample regime by fixing $N$ to be 100. We present the results here up to $\e = 0.5$, and the rest in the appendix.

Figure \ref{fig:FOREST_eps} shows that for small values of $\e$, QNO and S-QNO perform poorly, giving high FCR. S-QNO is able to resolve some of these issues by dividing the optimization into several steps, but still underperforms compared to our approach. SD gives a biased estimate of the bound for $\e$ significantly higher than $0$, as expected. Bootstrap gives valid bounds with low FCR only with near negligable values of $\e$, where the typical $\mmd$ estimate is approximately valid. 

The previous three experiments show that S-SD consistently gives credible and tight estimates of the upper and lower bounds on the value of the true $\mmd$. Next, we examine the sensitivity of S-SD to the number of steps $S$.

\subsection{Sensitivity to the choice of number of steps} \label{sec:step_sens}
\begin{wraptable}[16]{R}{0.45\textwidth}
\centering
\vspace{-1.5em}
\resizebox{0.45\textwidth}{!}{%
\begin{tabular}{c|ll}
\toprule
          & \multicolumn{2}{c}{S-SD (Ours)} \\
\cmidrule(r){2-3}
No. of Steps  & FCR    & MIW    \\
\cmidrule(r){2-3}
$2$   & $0.21 \pm (0.091$) & $0.082 \pm (0.001)$ \\
$3$ & $0.13 \pm (0.034)$ & $0.079 \pm (0.001)$ \\
$5$  & $0.0 \pm (0.0)$ & $0.088 \pm (0.001)$ \\
$10$  & $0.0 \pm (0.0)$ & $0.08 \pm (0.001)$ \\
$20$ & $0.0 \pm (0.0)$ & $0.091 \pm (0.001)$ \\
$50$ & $0.0 \pm (0.0)$ & $0.091 \pm (0.001)$ \\
\bottomrule
\end{tabular}
}
\caption{Varying number of steps for S-SD in FOREST $(N = 2000, d = 54)$ with $\e = 0.2$. Standard errors (in parentheses) over 100 trials. Results imply that setting $S$ to be large gives lower FCR.}
\label{tab:sensitivity_step}
\end{wraptable}

Here, we examine the sensitivity of S-SD to the number of steps $S$. To do so, we sample $n=2000$ from FOREST, vary the value of $S$, and examine the performance of our main approach, S-SD. 
We repeat the experiment 100 times using 100 different samples from FOREST, each of size 2000 to compute the standard errors around the FCR and MIW. 

Table ~\ref{tab:sensitivity_step} shows the results. The results imply that we can get bound estimates that give a FCR of zero even with a very few number of steps. The MIW increase slightly and starts to plateau as the number of steps increases. 
This implies that a reasonable choice of $S$ to ensure a low FCR would be the largest possible value which does not lead to a computationally prohibitive number of iterations. Recall that there is a natural upper bound on $S = m$.
In the appendix, we repeat this experiment for S-QNO showing similar robustness.

\section{Conclusion}
We studied the problem of comparing two distributions when the data is collected with some measurement error. Specifically, we showed that typical estimates of kernel based two-sample tests are unreliable when the data is measured with some $\e$ contamination, where an $\e$ proportion of one sample is erroneously included with the other. We showed both empirically and theoretically that the typical optimization approaches have an unfavorable dependence on the size of the contaminated set. Instead, we proposed a stepwise approach to estimate credible and tight upper and lower bounds and showed that it converges faster than alternatives to the true upper and lower bounds. Empirically, we showed that it outperforms all baselines. 
Looking beyond this work, it would be both interesting and important to study other commonly occurring measurement error mechanisms and study their effect on measuring the $\mmd$ and other related independence tests such as the Hilbert Schmidt independence criterion. In addition, one possible limitation of our work is the assumption that $\e$ is known \emph{a priori}. Future work addressing unreliable estimates of $\e$ represents an important future direction. 

\bibliography{bib}
\bibliographystyle{unsrtnat}
\newpage
\appendix
\section{Proof for proposition~\ref{prop:opt_main}}
Before proceeding to the main proof, we restate the following definition from ~\citet{gretton2012kernel}. 
\begin{thmappdef}
[Restated definition 30 in \cite{gretton2012kernel}]. Let $\cF$ be the unit ball in an RKHS, with kernel bounded according to $0 \leq k(x, y) \leq \kappa$. Let $Z$ be an i.i.d. sample of size $n$ drawn according to a probability measure $P_{Z}$ and let $\sigma_i$ be i.i.d and take values in $\{-1, 1\}$ with equal probability and $\boldsymbol{\sigma} = \{ \sigma_i\}_{i = 1}^n$. We define the Rademacher average:
\begin{align*}
    \cR_n(\cF, Z) = \E_{\boldsymbol{\sigma}} \sup_{f \in \cF} \Big| \frac{1}{n} \sum_i f(z_i) \Big| \leq \Big(\frac{\kappa}{n}\Big)^{1/2}
\end{align*}
\end{thmappdef}
\begin{thmappprop}
    [Restated Proposition~\ref{prop:opt_main} in the main text]     For $\mmd(P_{\overline{C}}, P_{X'}, P_{Y'})$ as defined in claim~\ref{clm:sharp}, $\widehat{C}_{\circ}$ as defined in equation~\ref{eq:optim_io}, with $\#(\widehat{C}_{\circ}) = m$, $0 \leq k(x', y') \leq \kappa$ for all $x', y' \in \cX', \cY'$, we have that: 
\begin{align*}
    P_{X', Y'} \bigg\{ | \mmd(P_{\overline{C}}, P_{X'}, P_{Y'}) - \mmd(\widehat{C}_{\circ}, X', Y') | > b_0 + \varepsilon \bigg\} \leq 2\exp\bigg(\frac{ - \varepsilon^2 n}{b_1}\bigg), 
\end{align*}
for $b_0 = 4\sqrt{\kappa} (n^{-1/2} + \e m)$ and $b_1 = 2 \kappa((1 - \e)(1 - \e + \e m)^2  + (1 + \e)  (1 + \e + \e m)^2)$
\end{thmappprop}

\begin{thmproof}
    Define $\hat{c}_i^\circ$ such that $\widehat{C}_{\circ} = \{\hat{c}_i^\circ \}_{i = 1}^m$ and consider the absolute difference term: 
\begin{align*}
    & |\mmd(P_{\overline{C}}, P_{X'}, P_{Y'}) - \mmd(\widehat{C}_{\circ}, X', Y') | \\
    & = \bigg| \sup_{f \in \cF} \Big[(1 - \e) \E_{P_{X'}}f(X') - (1 + \e) \E_{P_{Y'}}f(Y') + 2 \e \E_{\overline{C}}f(\overline{C}) \Big] \\
    & \quad -  \sup_{f \in \cF} \Big[\frac{(1 - \e)}{n} \sum_i f(x'_i) - \frac{(1 + \e)}{n} \sum_i f(y'_i) + \frac{2\e}{n} \sum_i f(\hat{c}^{\circ}_i) \Big] \bigg| \\
    & \leq \sup_{f \in \cF} \Big|(1 - \e) \E_{P_{X'}}f(X') - (1 + \e) \E_{P_{Y'}}f(Y') + 2 \e \E_{\overline{C}}f(\overline{C})  \\
    & \quad - \frac{(1 - \e)}{n} \sum_i f(x'_i) + \frac{(1 + \e)}{n} \sum_i f(y'_i) - \frac{2\e}{n} \sum_i f(\hat{c}^{\circ}_i))  \Big| \\
    & := \Delta(X', Y', P_{X'}, P_{Y'})
    \end{align*}

We will next attempt to bound the difference between $\Delta_{\cD}(P_{X'}, P_{Y'}, X', Y')$ and its expectation by applying McDiarmid's inequality. To do so, we first need to verify that $\Delta_{\cD}(P_{X'}, P_{Y'}, X', Y')$ satisfies the bounded difference property. We do so in two steps. In the first step, we consider the case where we replace one of the $X'$ samples. Specifically, we consider the data $\cD^{X'}_{\pi j} = \{X'_{\pi j}, Y'\}$, where $X'_{\pi j} = \{x'_1, x'_2, \hdots, x'_{i - 1}, x'_j, x'_{i+1}, \hdots x'_{(1 - \e)n} \}$. 
Let $\widetilde{C}_{\circ}$ denote the estimate of $\widehat{C}$ according to equation~\ref{eq:optim_io} using $\cD^{X'}_{\pi j}$ rather than $\cD$. In that case, we have that:  
\begin{align}\label{eq:prop_opt_replacex}
    \nonumber & | \Delta_{\cD}(P_{X'}, P_{Y'}, X', Y')  - \Delta_{\cD^{X'}_j}(P_{X'}, P_{Y'}, X'_{\pi j}, Y')| \\
    \nonumber & \leq \sup_{f} \Big| \frac{(1 - \e)}{n} (\sum_i f(x'_i) -f(x'_i) + f(x'_j)) - \frac{(1 + \e)}{n} \sum_i f(y'_i) \\
    \nonumber & \quad + \frac{2\e}{n} \sum_i f(\tilde{c}^{\circ}_i) 
    - \frac{(1 - \e)}{n} \sum_i f(x'_i) + \frac{(1 + \e)}{n} \sum_i f(y'_i) - \frac{2\e}{n} \sum_i f(\hat{c}^{\circ}_i) \Big|\\
    \nonumber & \leq \sup_{f} \Big| \frac{(1 - \e)}{n} (- f(x'_i)  + f(x'_j)) 
    + \frac{2\e}{n} \sum_i f(\tilde{c}^{\circ}_i) 
    - \frac{2\e}{n} \sum_i f(\hat{c}^{\circ}_i) \Big| \\
    \nonumber & \leq \frac{(1 - \e)}{n} (\sup_f |f(x_i')| + \sup_f |f(x_j')|) + \frac{2 \e}{n} \sup_f (\sum_i f(\tilde{c}^{\circ}_i) - \sum_i f(\hat{c}^{\circ}_i)) \\
    & \leq \frac{(1 - \e)}{n} (2\sqrt{\kappa}) + \frac{2 \e}{n} (m \sqrt{\kappa}) = \frac{2 \sqrt{k}}{n} (1 - \e + \e m)
\end{align}

Second, we consider the case where we replace one of the $Y'$ samples. Specifically, we consider the data $\cD^{Y'}_{\pi j} = \{X', Y'_{\pi j}\}$, where $Y'_{\pi j} = \{y'_1, y'_2, \hdots, y'_{i - 1}, y'_j, y'_{i+1}, \hdots y'_{(1 + \e)n} \}$. In that case, by a similar construction to the previous case, we have that: 
\begin{align}\label{eq:prop_opt_replacey}
    & | \Delta_{\cD}(P_{X'}, P_{Y'}, X', Y')  - \Delta_{\cD^{Y'}_j}(P_{X'}, P_{Y'}, X', Y'_{\pi j})| 
    \leq \frac{2 \sqrt{k}}{n} (1 + \e + \e m)
\end{align}

Combining the results from equations~\ref{eq:prop_opt_replacex} and~\ref{eq:prop_opt_replacey}, we can apply McDiarmid with denominator: 
\begin{align*}
    & (1 - \e)n \Big(\frac{2 \sqrt{k}}{n} (1 - \e + \e m)\Big)^2 + (1 + \e)n \Big(\frac{2 \sqrt{k}}{n} (1 + \e + \e m)\Big)^2 \\
    & = \frac{4 \kappa}{n}\Big((1 - \e)(1 - \e + \e m)^2  + (1 + \e)  (1 + \e + \e m)^2\Big).
\end{align*}
I.e.,:
\begin{align}\label{prop:opt_unbounded_e}
    & P_{X', Y'}\bigg\{ \Delta_{\cD}(P_{X'}, P_{Y'}, X', Y') - \E_{X', Y'}\Big[ \Delta_{\cD}(P_{X'}, P_{Y'}, X', Y')\Big] > \varepsilon \bigg\} 
     \leq 2 \exp\bigg(\frac{ - \varepsilon^2 n}{b_1}\bigg),
\end{align}
where $b_1 = 2 \kappa((1 - \e)(1 - \e + \e m)^2  + (1 + \e)  (1 + \e + \e m)^2)$.

It remains to control $\E_{X', Y'}\Big[ \Delta_{\cD}(P_{X'}, P_{Y'}, X', Y')\Big] $. To do so we use the $\beta$-stability property and  symmetrization \cite{van1996weak}. 
We note that the $\beta$-stability of the hypothesis is a direct consequence of the boundedness of $k(., .)$ by $\kappa$.   
Let $\dr{X}$ and $\dr{Y}$ be i.i.d samples of sizes $(1-\e) n$ and $(1 + \e) n$ respectively, we have that: 

\begin{align*}
& \E_{X', Y'}\Big[ \Delta_{\cD}(P_{X'}, P_{Y'}, X', Y')\Big] \\
& = \E_{X', Y'}  \sup_{f} \Big|(1 - \e) \E_{P_{X'}}f(X') - (1 + \e) \E_{P_{Y'}}f(Y') + 2 \e \E_{\overline{C}}f(\overline{C})  \\
& \quad - \frac{(1 - \e)}{n} \sum_i f(x'_i) + \frac{(1 + \e)}{n} \sum_i f(y'_i) - \frac{2\e}{n} \sum_i f(\hat{c}^{\circ}_i))  \Big| \\
& = \E_{X', Y'} \sup_{f} \Big| (1 - \e) \E_{\dr{X}}\bigg(\frac{1}{n} \sum_i f(\dr{x}_i) \bigg)  - \frac{1 - \e}{n} \sum_i f(x'_i) - (1 + \e) \E_{\dr{Y}}\bigg( \frac{1}{n} f(\dr{y}_i)\bigg)+ \frac{1 + \e}{n} \sum_i f(y'_i) \\
& \quad  + 2 \e \E_{\dr{X}, \dr{Y}}\bigg( \frac{1}{n}  f(\dot{c}^{\circ}_i) \bigg)  - \frac{2 \e}{n} \sum_i f(\hat{c}^{\circ}_i) \Big| \\
& \leq \E_{X', Y', \dr{X}, \dr{Y}} \sup_f \Big|  \frac{1 - \e}{n} \sum_i f(\dr{x}_i) - \frac{1 - \e}{n} \sum_i f(x'_i) 
-  \frac{1 + \e}{n} \sum_i f(\dr{y}_i) + \frac{1 + \e}{n} \sum_i f(y'_i) \\
& \quad  +  \frac{2 \e}{n} \sum_i f(\dot{c}^{\circ}_i)- \frac{2 \e}{n} \sum_i f(\hat{c}^{\circ}_i) \Big| \\
& \leq \E_{X', Y', \dr{X}, \dr{Y}} \sup_{f} \Big|  \frac{1 - \e}{n} \sum_i f(\dr{x}_i) - \frac{1 - \e}{n} \sum_i f(x'_i) 
-  \frac{1 + \e}{n} \sum_i f(\dr{y}_i) + \frac{1 + \e}{n} \sum_i f(y'_i) \Big|\\
& \quad  +  \E_{X', Y', \dr{X}, \dr{Y}} \sup_{f} \Big| \frac{2 \e}{n} \sum_i f(\dot{c}^{\circ}_i)- \frac{2 \e}{n} \sum_i f(\hat{c}^{\circ}_i) \Big| \\
& \leq \E_{X', Y', \dr{X}, \dr{Y}, \sigma', \dr{\sigma}} \sup_f \Big|  \frac{1 - \e}{n} \sum_i \sigma'_i (f(\dr{x}_i) - f(x'_i) )
+  \frac{1 + \e}{n} \sum_i \dr{\sigma}_i (f(\dr{y}_i) - f(y'_i) )\Big| \\
& + \sup_{X', Y',  \dr{X}, \dr{Y}} \Big| \frac{2 \e}{n} \sum_i f(\dot{c}^{\circ}_i)- \frac{2 \e}{n} \sum_i f(\hat{c}^{\circ}_i) \Big| \\
& \leq \E_{X', \dr{X}, \sigma} \sup_{f} \bigg|  \frac{1 - \e}{n} \sum_i \sigma'_i (f(\dr{x}_i) - f(x'_i) ) \Big| + \E_{Y', \dr{Y}, \sigma}  \sup_{f} \Big|  \frac{1 + \e}{n} \sum_i \dr{\sigma}_i (f(\dr{y}_i) - f(y'_i) ) \Big|  \\
& + \quad \frac{2 \e}{n} \sup_{X', Y',  \dr{X}, \dr{Y}} \Big| \sum_i f(\dot{c}^{\circ}_i)- \sum_i f(\hat{c}^{\circ}_i) \Big| \\
& \leq 2 [(1 - \e)\cR_n(\cF, X') + (1 + \e) \cR_n(\cF,Y')] + 2\e m \sqrt{\kappa}] \\
& \leq 2 [(1-\e) (\kappa /n)^{1/2} + (1+\e) (\kappa /n)^{1/2} + 2\e m \kappa^{1/2}] \\
& \leq 4\sqrt{\kappa} (n^{-1/2} + \e m).
\end{align*}
Substituting $4\sqrt{\kappa} (n^{-1/2} + \e m)$ for $\E_{X', Y'}\Big[ \Delta_{\cD}(P_{X'}, P_{Y'}, X', Y')\Big]$ in~ equation \ref{prop:opt_unbounded_e} gives the desired result.
\end{thmproof}

\section{Proof for proposition~\ref{prop:sd_asymp}}
Before stating the main proof, we begin by outlining the following definition, and lemmas. 
\begin{thmappdef}
    Random variable $Z$ has first-order stochastic dominance (or stochastic dominance for short) over random variable $Z'$ if for any outcome $t$, $Z$ gives at least as high a probability of receiving at least $t$ as does $Z'$, and for some $t$, $Z$ gives a higher probability of receiving at least $t$. 
\end{thmappdef}

\begin{thmapplem}\label{lem:decompose}
Let $(\cY', \Omega)$ be a measurable space with $Y' \in \cY'$, and let $\cP$ be all the probability distributions on $(\cY', \Omega)$. For $\cP(\alpha) = \{ (P_{Y'}(Y') - (1 - \alpha) \varphi)/\alpha : \varphi \in \cP \}$. 
We have that 
\begin{align*}
    {\arg\sup}_{P_C \in \cP(\alpha)}\mmd(P_C,  P_{X'}, P_{Y'}) = {\arg \sup}_{P_C \in \cP(\alpha)} \E_{P_C}[\tilde{f}'(C)],  
\end{align*}
where 
\begin{align}\label{eq:pop_f_to_opt}
    \tilde{f}'(C) = (1 - \e) \E_{P_{X'}}[k(C, X')] - (1 + \e) \E_{P_{Y'}}[k(C, Y')] + \e \E_{P_C} k(C, C)
\end{align}
\end{thmapplem}

\begin{thmproof}
    The proof is a straight forward derivation from the definition of the $\mmd$ and the witness function. 
    We present the derivation below, with all $\sup_{P_C}$ to be understood as $\sup_{P_C \in \cP(\alpha)}$.
    We use $\widetilde{X}$ to denote $X' \cup C$ and $\widetilde{Y}$ to denote $Y' \setminus C$ for an arbitrary $C$. 

    \begin{align*}
        & {\arg\sup}_{P_C} \Big[\mmd(P_C,  P_{X'}, P_{Y'})\Big] \\
        & = {\arg\sup}_{P_C} \Bigg[\sup_{f \in \cF} \Big[ \E_{P_{\widetilde{X}}}[f(\widetilde{X})] - \E_{P_{\widetilde{Y}}}[f(\widetilde{Y})]\Big]\Bigg]\\
        & = {\arg\sup}_{P_C} \Big[\E_{P_{\widetilde{X}}}[k(\widetilde{X}, \widetilde{X})] - 
        \E_{P_{\widetilde{X}}}\E_{P_{\widetilde{Y}}}[k(\widetilde{X}, \widetilde{Y})] -
        \E_{P_{\widetilde{X}}}\E_{P_{\widetilde{Y}}}[k(\widetilde{X}, \widetilde{Y})] + 
        \E_{P_{\widetilde{Y}}}[k(\widetilde{Y}, \widetilde{Y})]\Big] \\
        & = {\arg\sup}_{P_C}\Big[(1 - \e)^2 \E_{P_{X'}}[k(X', X')] + (1 + \e)^2 \E_{P_{y'}}[k(y', y')] \\
        & \qquad - 2(1+\e)(1-\e) \E_{P_{X'}}\E_{P_{Y'}}[k(X', Y')] 
        + 4\e \big( (1-\e) \E_{P_C}\E_{P_{X'}}[k(C, X')] \\
        & \qquad - (1 + \e) \E_{P_C}\E_{P_{Y'}}[k(C, Y')]+ \E_{P_C}\E_{P_{C}}[k(C,C)]\Big] \\
        & = {\arg\sup}_{P_C}\Big[\E_{P_C}\big[(1-\e)\E_{P_{X'}}[k(C, X')] - (1 + \e) \E_{P_{Y'}}[k(C, Y')]+\E_{P_{C}}[k(C,C)]\big]\Big] \\
        & = {\arg\sup}_{P_C} \big[\tilde{f}'(C)], 
    \end{align*}
    which completes the proof.
\end{thmproof}

Note that the empirical version of equation~\ref{eq:pop_f_to_opt} corresponds to equation~\ref{eq:optim} in the main text. 

\begin{thmappcol}\label{col:decompose}
Under the same conditions as Lemma~\ref{lem:decompose}, and for a sufficiently small $\e$, we have that 
\begin{align*}
    {\arg\sup}_{P_C \in \cP(\alpha)}\mmd(P_C,  P_{X'}, P_{Y'}) \lessapprox {\arg \sup}_{P_C \in \cP(\alpha)} \E_{P_C}[f'(C)],  
\end{align*}
where 
\begin{align*}
    f'(C) = (1 - \e) \E_{P_{X'}}[k(C, X')] - (1 + \e) \E_{P_{Y'}}[k(C, Y')]
\end{align*}
\end{thmappcol}
\begin{thmproof}
    The proof directly follows from Lemma~\ref{lem:decompose} and the fact that for a sufficiently small $\e$, we have that $f'(C) \approx \tilde{f}'(C)$. 
\end{thmproof}

\begin{thmappprop}
    [Restated proposition~\ref{prop:sd_asymp} from the main text] Let $C_{{\gamma}}$ be the solution to equation~\ref{eq:sd_optim} as $n \rightarrow \infty$. For a sufficiently small $\e$, we have that $P_{C_{\gamma}}  = P_{\overline{C}}$, where $P_{\overline{C}}$ is defined as the distribution that maximizes the third term in claim~\ref{clm:sharp}.
\end{thmappprop}

\begin{thmproof}
    Recall that: 
    \begin{align*}
       P_{Y'}(Y')  = (1 - \alpha)P_{Y}(Y) + \alpha P_{C^*}(C^*), 
    \end{align*}
    and note that the kernel $k$ is a measurable mapping, hence $f'$ is also a measurable mapping. 
    This implies that $f'(Y')$ is measurable with respect to $Y'$ and we can express the distribution over $f'(Y')$. 
    Letting $Q_{Y'} := P_{Y'}(f'(Y'))$, $Q_{Y} := P_{Y}(f'(Y))$, and $Q_{C^*} := P_{C^*}(f(C^*))$, we have that: 
    \begin{align*}
        Q_{Y'}(Y')  = (1 - \alpha)Q_{Y}(Y) + \alpha Q_{C^*}(C^*).
    \end{align*}
    Using the notation $Q_{Y'}[-\infty, t]$ to denote the cumulative distribution function (CDF) of $Q_{Y'}(Y')$ from values $-\infty$ to $t$, we can write the CDF over $C_\gamma$ as the CDF of a truncated distribution, which gives us the following: 
    \begin{align*}
        Q_{C_\gamma}[-\infty, t] = 
        \begin{cases}
            0 & \text{if } t < \gamma \\
           \Big(Q_{Y'}[-\infty, t] - (1 - \alpha) \Big)/\alpha & \text{if } t \ge \gamma.
        \end{cases}
    \end{align*}

    Consider the following distribution: 
    \begin{align*}
        \varphi_0[-\infty, t] = 
        \begin{cases}
           Q_{Y'}[-\infty, t]/(1 - \alpha)  & \text{if } t < \gamma \\
           1 & \text{if } t \ge \gamma.
        \end{cases}
    \end{align*}
Note that:
\begin{align*}
    (1 - \alpha) \varphi_0[-\infty, t] + \alpha Q_{C_\gamma}[-\infty, t] = Q_{Y'}[-\infty, 1] 
\end{align*}
which means that $Q_{C_\gamma} \in \cP(\alpha)$. 
Next we will make the argument that $Q_{C_\gamma}$ stochastically dominates all other distributions in $\cP(\alpha)$. 
Note that for any $\varphi_1$, if $t < \gamma$
\begin{align*}
    Q_{C_\gamma}[-\infty, t] - \varphi_1[-\infty, t]  = 0 - \varphi_1[-\infty, t]  \leq 0. 
\end{align*}
However, suppose that there exists some $\varphi_1 \in \cP(\alpha)$, and that it stochastically dominates $Q_{C_\gamma}$. I.e., for $t \geq \gamma$: 
\begin{align*}
     \varphi_1[-\infty, t] & < Q_{C_\gamma}[-\infty, t]\\
    \Rightarrow  \varphi_1[-\infty, t] & < \Big(Q_{Y'}[-\infty, t] - (1 - \alpha) \Big)/\alpha\\
    \Rightarrow  \alpha \varphi_1[-\infty, t] & < Q_{Y'}[-\infty, t] - (1 - \alpha),
\end{align*}
Hence we have that $(1 - \alpha) \varphi + \alpha \varphi_1 < Q_{Y'}[-\infty, 1]$ for all $\varphi \in \cP$, which implies that $\varphi_1 \not \in \cP(\alpha)$, which is a contradiction. 

This shows that $Q_{C_\gamma}[-\infty, t]$ stochastically dominates all distributions in $\cP(\alpha)$, which means that: 
\begin{align*}
    \E_{Q_{C_\gamma}}[f'(C_{\gamma})] & > \E_{Q_{C}}[f'(C)]\\
    \Rightarrow \E_{P_{C_\gamma}}[f'(C_{\gamma})] & > \E_{P_{C}}[f'(C)]
\end{align*}
for all $P_C \not= P_{C_\gamma}$. Since $\E_{P_{\overline{C}}}[f'(\overline{C})] > \E_{P_{C}}[f'(C)]$ for all $P_C \not= P_{\overline{C}}$, and by Corollary~\ref{col:decompose}, we have that $ \E_{P_{C_\gamma}}[f'(C_{\gamma})] = \E_{P_{\overline{C}}}[f'(\overline{C})]$, which completes the proof. 
\end{thmproof}


\section{Proof for proposition~\ref{prop:sd_main}}
\begin{thmappprop} [Restated proposition ~\ref{prop:sd_main} in main text]
    For $\mmd(P_{\overline{C}}, P_{X'}, P_{Y'})$ as defined in claim~\ref{clm:sharp}, $\widehat{C}_{\hat{\gamma}}$ as defined in equation~\ref{eq:sd_optim} and $\kappa$ such that $0 \leq k(x, y) \leq \kappa$ for all $x, y \in \cX$. Then as for a sufficiently small $\e$: 
    \begin{align*}
        P_{X', Y'} \bigg\{ | \mmd(P_{\overline{C}}, P_{X'}, P_{Y'}) - \mmd(\widehat{C}_{\hat{\gamma}}, X', Y') | > b_0 + \varepsilon \bigg\} \leq 2\exp\bigg(\frac{ - \varepsilon^2 n}{b_1}\bigg)
    \end{align*}
for $b_0 = 4( \kappa /n)^{\sfrac{1}{2}}(1 + \e)$ and $b_1 = 2 \kappa \big((1 -\e)^3 + (1+\e) (1 + 3 \e)^2\big)$
\end{thmappprop}

\begin{thmproof}
Consider the absolute difference term
\begin{align*}
    & |\mmd(P_{\overline{C}}, P_{X'}, P_{Y'}) - \mmd(\widehat{C}_{\hat{\gamma}}, X', Y') | \\
    & = \bigg| \sup_{f} \Big[(1 - \e) \E_{P_{X'}}f(X') - (1 + \e) \E_{P_{Y'}}f(Y') + 2 \e \E_{\overline{C}}f(\overline{C}) \Big] \\
    & \quad -  \sup_{f} \Big[\frac{(1 - \e)}{n} \sum_i f(x'_i) - \frac{(1 + \e)}{n} \sum_i f(y'_i) + \frac{2\e}{n} \sum_i f(\hat{c}^{\hat{\gamma}}_i) \Big] \bigg| \\
    & \leq \sup_{f} \Big|(1 - \e) \E_{P_{X'}}f(X') - (1 + \e) \E_{P_{Y'}}f(Y') + 2 \e \E_{\overline{C}}f(\overline{C})  \\
    & \quad - \frac{(1 - \e)}{n} \sum_i f(x'_i) + \frac{(1 + \e)}{n} \sum_i f(y'_i) - \frac{2\e}{n} \sum_i f(\hat{c}^{\hat{\gamma}}_i)  \Big| \\
    & =  \sup_{f} \Big| (1 - \e) \E_{P_{X'}}f(X') - (1 + \e) \E_{P_{Y'}}f(Y') + 2 \e \E_{\overline{C}}f(\overline{C})  \\
    & \quad - \frac{(1 - \e)}{n} \sum_i f(x'_i) + \frac{(1 + \e)}{n} \sum_i f(y'_i) - \frac{2 \e}{n} \sum_i \indic \{ f(y'_i) \geq \hat{\gamma}\} f(y'_i) \Big| \\
    & := \Delta_{\cD}(P_{X'}, P_{Y'}, X', Y') 
\end{align*}

We will next attempt to bound the difference between $\Delta_{\cD}(P_{X'}, P_{Y'}, X', Y')$ and its expectation by applying McDiarmid's inequality. To do so, we first need to verify that $\Delta_{\cD}(P_{X'}, P_{Y'}, X', Y')$ satisfies the bounded difference property. We do so in two steps. In the first step, we consider the case where we replace one of the $X'$ samples. Specifically, we consider the data $\cD^{X'}_{\pi j} = \{X'_{\pi j}, Y'\}$, where $X'_{\pi j} = \{x'_1, x'_2, \hdots, x'_{i - 1}, x'_j, x'_{i+1}, \hdots x'_{(1 - \e)n} \}$. In that case, we have that: 

\begin{align}
    \nonumber & | \Delta_{\cD}(P_{X'}, P_{Y'}, X', Y')  - \Delta_{\cD^{X'}_j}(P_{X'}, P_{Y'}, X'_{\pi j}, Y')| \\
    \nonumber & =  \sup_{f} \Big| (1 - \e) \E_{P_{X'}}f(X') - (1 + \e) \E_{P_{Y'}}f(Y') + 2 \e \E_{\overline{C}}f(\overline{C})  \\
    \nonumber & \quad - \frac{(1 - \e)}{n} \sum_i f(x'_i) + \frac{(1 + \e)}{n} \sum_i f(y'_i) - 2 \e \frac{1}{n} \sum_i \indic \{ f(y'_i) \geq \hat{\gamma}\} f(y'_i) \Big| \\
    \nonumber & \quad - \sup_{f} \Big| (1 - \e) \E_{P_{X'}}f(X') - (1 + \e) \E_{P_{Y'}}f(Y') + 2 \e \E_{\overline{C}}f(\overline{C})  \\
    \nonumber & \quad - \frac{(1 - \e)}{n} \sum_i f(x'_i) + \frac{(1 + \e)}{n} \sum_i f(y'_i) - 2 \e \frac{1}{n} \sum_i \indic \{ f(y'_i) \geq \tilde{\gamma}\} f(y'_i) + \frac{1 - \e}{n} ( f(x'_j) - f(x'_i)) \Big| \\
     \nonumber & \leq \sup_{f, \gamma} \Big| (1 - \e) \E_{P_{X'}}f(X') - (1 + \e) \E_{P_{Y'}}f(Y') + 2 \e \E_{\overline{C}}f(\overline{C})  \\
    \nonumber & \quad - \frac{(1 - \e)}{n} \sum_i f(x'_i) + \frac{(1 + \e)}{n} \sum_i f(y'_i) - 2 \e \frac{1}{n} \sum_i \indic \{ f(y'_i) \geq \gamma\} f(y'_i) \Big| \\
    \nonumber & \quad - \sup_{f} \Big| (1 - \e) \E_{P_{X'}}f(X') - (1 + \e) \E_{P_{Y'}}f(Y') + 2 \e \E_{\overline{C}}f(\overline{C})  \\
    \nonumber & \quad - \frac{(1 - \e)}{n} \sum_i f(x'_i) + \frac{(1 + \e)}{n} \sum_i f(y'_i) - 2 \e \frac{1}{n} \sum_i \indic \{ f(y'_i) \geq \gamma\} f(y'_i) + \frac{1 - \e}{n} ( f(x'_j) - f(x'_i)) \Big| \\
    \nonumber & \leq \frac{1 - \e}{n} \sup_{f} \Big|( f(x'_i) - f(x'_j)) \Big| \\
    \nonumber & \leq \frac{1 - \e}{n} \Big(\sup_{f} |( f(x'_i)| + \sup_f |f(x'_j))|\Big) \\
    & \label{eq:prop_sd_replacex} \leq \frac{2 (1 - \e)}{n} \sqrt{\kappa} 
\end{align}

Second, we consider the case where we replace one of the $Y'$ samples. Specifically, we consider the data $\cD^{Y'}_{\pi j} = \{X', Y'_{\pi j}\}$, where $Y'_{\pi j} = \{y'_1, y'_2, \hdots, y'_{i - 1}, y'_j, y'_{i+1}, \hdots y'_{(1 + \e)n} \}$. In that case, we have that: 

\begin{align}
    \nonumber & | \Delta_{\cD}(P_{X'}, P_{Y'}, X', Y')  - \Delta_{\cD^{Y'}_j}(P_{X'}, P_{Y'}, X', Y'_{\pi j})| \\
    \nonumber & \leq  \sup_{f} \Big| (1 - \e) \E_{P_{X'}}f(X') - (1 + \e) \E_{P_{Y'}}f(Y') + 2 \e \E_{\overline{C}}f(\overline{C})  \\
    \nonumber & \quad - \frac{(1 - \e)}{n} \sum_i f(x'_i) + \frac{(1 + \e)}{n} \sum_i f(y'_i) - 2 \e \frac{1}{n} \sum_i \indic \{ f(y'_i) \geq \hat{\gamma}\} f(y'_i) \Big| \\
    \nonumber & \quad - \sup_{f} \Big| (1 - \e) \E_{P_{X'}}f(X') - (1 + \e) \E_{P_{Y'}}f(Y') + 2 \e \E_{\overline{C}}f(\overline{C})  \\
    \nonumber & \quad - \frac{(1 - \e)}{n} \sum_i f(x'_i) + \frac{(1 + \e)}{n} \sum_i f(y'_i) - 2 \e \frac{1}{n} \sum_i \indic \{ f(y'_i) \geq \tilde{\gamma}\} f(y'_i) \\
    \nonumber & \quad - \frac{1 + \e}{n} ( f(y'_i) - f(y'_j)) + \frac{2 \e}{n}( \indic \{ f(y'_i) \geq \hat{\gamma}\}f(y'_i) - \indic \{ f(y'_j) \geq \tilde{\gamma}\}f(y'_j)) \Big| \\
    \nonumber & = \sup_{f} \Big| - \frac{1 + \e}{n} ( f(y'_i) - f(y'_j)) + \frac{2 \e}{n}(\indic \{ f(y'_i) \geq \hat{\gamma}\}f(y'_i) - \indic \{ f(y'_j) \geq \tilde{\gamma}\}f(y'_j)) \Big| \\
    \nonumber & \leq  \frac{1 + \e}{n}  \sup_{f}\Big|( f(y'_i) - f(y'_j)) \Big| + \frac{2 \e}{n} \sup_f \Big|(\indic \{ f(y'_i) \geq \hat{\gamma}\}f(y'_i) - \indic \{ f(y'_j) \geq \tilde{\gamma}\}f(y'_j)) \Big| \\
    \nonumber & \leq  \frac{1 + \e}{n}  \sup_{f}\Big|( f(y'_i) - f(y'_j)) \Big| + \frac{2 \e}{n} \sup_f | f(y'_i) - f(y'_j) \Big| \\
    \nonumber & \leq  \frac{1 + \e}{n}  \Big(\sup_{f}|( f(y'_i)| + \sup_f |f(y'_j))|\Big) + \frac{2 \e}{n} \Big(\sup_f |f(y'_i)| + \sup_f |f(y'_j)| \Big) \\
    & \leq \frac{2 (1 + \e)}{n} \sqrt{\kappa} + \frac{4 \e}{n} \sqrt{\kappa} = \frac{2 \sqrt{\kappa}}{n} (1 + 3 \e) \label{eq:prop_sd_replacey}
\end{align}
Combining the results from equations~\ref{eq:prop_sd_replacex} and~\ref{eq:prop_sd_replacey}, we get that we can apply McDiarmid with the following denominator: 
\begin{align*}
    (1 - \e) n \Big(\frac{2 (1 - \e)}{n} \sqrt{\kappa}\Big)^2 + (1 + \e) n \Big(\frac{2 \sqrt{\kappa}}{n} (1 + 3 \e)\Big)^2 = \frac{4\kappa}{n} \Big((1 -\e)^3 + (1+\e) (1 + 3 \e)^2\Big), 
\end{align*}
to obtain
\begin{align}\label{eq:prop_sd_uncontrolled_E}
    & P_{X', Y'}\bigg\{ \Delta_{\cD}(P_{X'}, P_{Y'}, X', Y') - \E_{X', Y'}\Big[ \Delta_{\cD}(P_{X'}, P_{Y'}, X', Y')\Big] > \varepsilon \bigg\} \\ 
    & \leq 2 \exp\bigg(\frac{ - \varepsilon^2 n}{2 \kappa \big((1 -\e)^3 + (1+\e) (1 + 3 \e)^2\big)}\bigg).
\end{align}

Next, we seek to control the expectation, $\E_{X', Y'}\Big[ \Delta_{\cD}(P_{X'}, P_{Y'}, X', Y')\Big]$. To do so we use symmetrization \cite{van1996weak}. Let $\dr{X}$ and $\dr{Y}$ be i.i.d samples of sizes $(1-\e) n$ and $(1 + \e) n$ respectively, we have that: 
\begin{align*}
& \E_{X', Y'}\Big[ \Delta_{\cD}(P_{X'}, P_{Y'}, X', Y')\Big] \\
& = \E_{X', Y'} \sup_{f} \Big| (1 - \e) \E_{P_{X'}}f(X')  - \frac{1 - \e}{n} \sum_i f(x'_i) - (1 + \e) \E_{P_{Y'}}f(Y') + \frac{1 + \e}{n} \sum_i f(y'_i) \\
    & \quad  + 2 \e \E_{\overline{C}}f(\overline{C})  - \frac{2 \e}{n} \sum_i \indic \{ f(y'_i) \geq \hat{\gamma}\} f(y'_i) \Big| \\
& = \E_{X', Y'} \sup_{f} \Big| (1 - \e) \E_{\dr{X}}\bigg(\frac{1}{n} \sum_i f(\dr{x}_i) \bigg)  - \frac{1 - \e}{n} \sum_i f(x'_i) \\
& \quad - (1 + \e) \E_{\dr{Y}}\bigg( \frac{1}{n} f(\dr{y}_i)\bigg)+ \frac{1 + \e}{n} \sum_i f(y'_i) \\
 & \quad  + 2 \e \E_{\dr{Y}}\bigg( \frac{1}{n} \indic \{ f(\dr{y}_i) \geq \dr{\gamma}\} f(\dr{y}_i) \bigg)  - \frac{2 \e}{n} \sum_i \indic \{ f(y'_i) \geq \hat{\gamma}\} f(y'_i) \Big| \\
& \leq \E_{X', Y', \dr{X}, \dr{Y}} \sup_{f} \Big|  \frac{1 - \e}{n} \sum_i f(\dr{x}_i) - \frac{1 - \e}{n} \sum_i f(x'_i) 
-  \frac{1 + \e}{n} \sum_i f(\dr{y}_i) + \frac{1 + \e}{n} \sum_i f(y'_i) \\
& \quad  +  \frac{2 \e}{n} \sum_i \indic \{ f(\dr{y}_i) \geq \dr{\gamma}\} f(\dr{y}_i)  - \frac{2 \e}{n} \sum_i \indic \{ f(y'_i) \geq \hat{\gamma}\} f(y'_i) \Big| \\
& \leq \E_{X', Y', \dr{X}, \dr{Y}} \sup_{f, \gamma} \Big|  \frac{1 - \e}{n} \sum_i f(\dr{x}_i) - \frac{1 - \e}{n} \sum_i f(x'_i) 
-  \frac{1 + \e}{n} \sum_i f(\dr{y}_i) + \frac{1 + \e}{n} \sum_i f(y'_i) \\
& \quad  +  \frac{2 \e}{n} \sum_i \indic \{ f(\dr{y}_i) \geq \gamma\} f(\dr{y}_i)  - \frac{2 \e}{n} \sum_i \indic \{ f(y'_i) \geq \gamma\} f(y'_i) \Big| \\
& \leq \E_{X', Y', \dr{X}, \dr{Y}, \sigma', \dr{\sigma}} \sup_{f, \gamma} \Big|  \frac{1 - \e}{n} \sum_i \sigma'_i (f(\dr{x}_i) - f(x'_i) )
+  \frac{1 + \e}{n} \sum_i \dr{\sigma}_i (f(\dr{y}_i) - f(y'_i) ) \\
    & \quad  +  \frac{2 \e}{n} \sum_{y'_i, \dr{y}_i \geq \gamma} \dr{\sigma}_i (f(\dr{y}_i) - f(y'_i)) \Big| \\
& \leq \E_{X', \dr{X}, \sigma} \sup_{f, \gamma} \bigg|  \frac{1 - \e}{n} \sum_i \sigma'_i (f(\dr{x}_i) - f(x'_i) ) \Big| + \E_{Y', \dr{Y}, \sigma}  \sup_{f, \gamma} \Big|  \frac{1 + \e}{n} \sum_i \dr{\sigma}_i (f(\dr{y}_i) - f(y'_i) ) \Big|\\
& \quad + \E_{Y', \dr{Y}, \sigma}  \sup_{f, \gamma} \Big|  \frac{2 \e}{n} \sum_{y'_i, \dr{y}_i \geq \gamma} \dr{\sigma}_i (f(\dr{y}_i) - f(y'_i))\Big| \\
& \leq 2 [(1 - \e)\cR_n(\cF, X') + (1 + \e) \cR_n(\cF, Y') + 2\e \cR_n(\cF, Y')] \\
& \leq 2[(1 - \e) \Big( \frac{\kappa}{n}\Big)^{\sfrac{1}{2}} + (1 + 3\e) \Big( \frac{\kappa}{n}\Big)^{\sfrac{1}{2}}] 
= 4 \Big( \frac{\kappa}{n}\Big)^{\sfrac{1}{2}} (1 + \e)
\end{align*}
Substituting $4 \Big( \frac{\kappa}{n}\Big)^{\sfrac{1}{2}} (1 + \e)$ in equation~\ref{eq:prop_sd_uncontrolled_E} yields the desired result. 
\end{thmproof}

\section{Additional results from the nonrandom contamination setting}

We show results presenting the typical estimate of the $\mmd$ assuming no contamination. We also reproduce the main results in sections \ref{sec:approach} in the MIMIC setting. We additionally include the same experiment as in table \ref{nonrandom_main_table} for $N=2000$.

\begin{figure}[H]
    \centering
    \includegraphics[width=\textwidth]{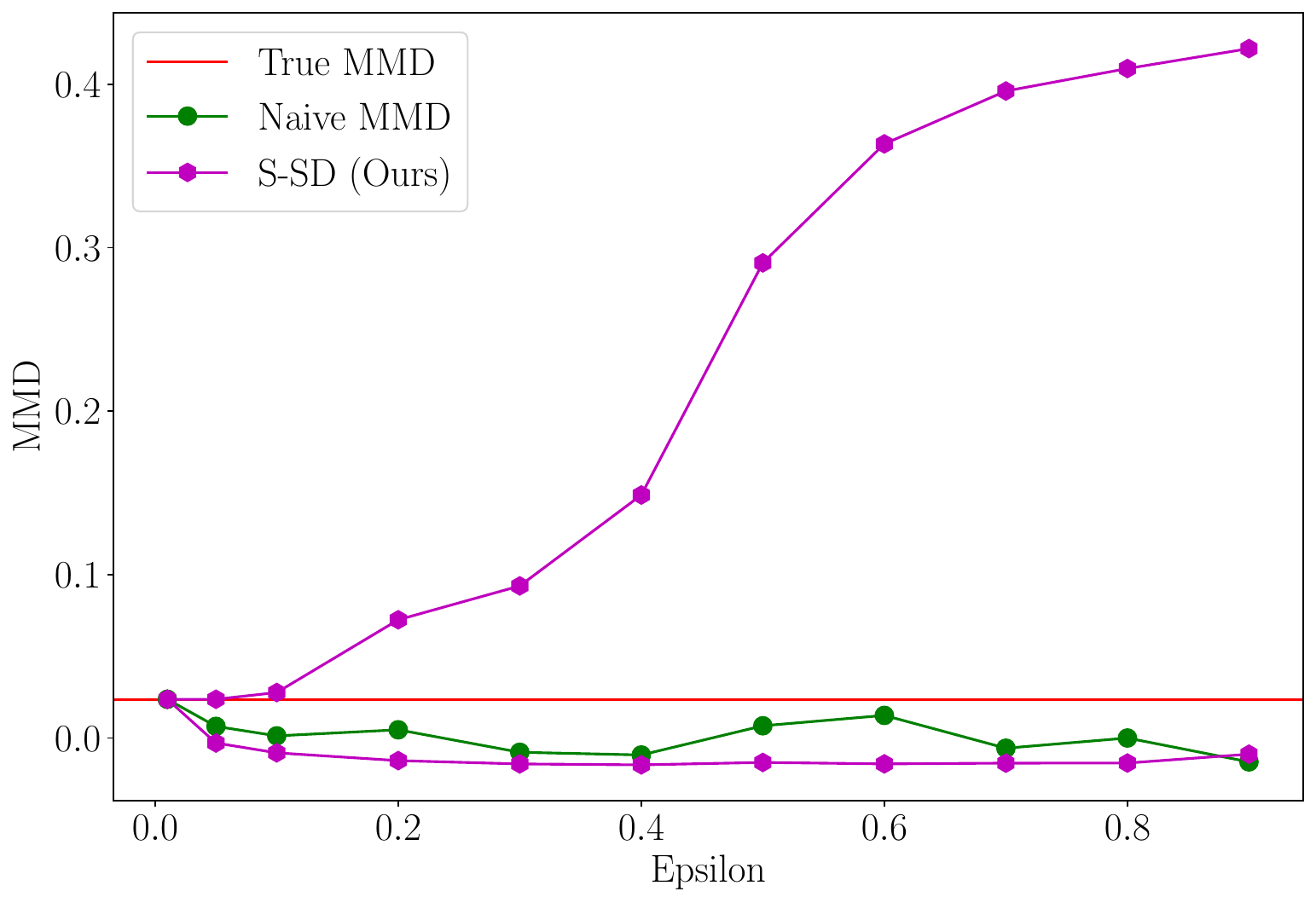}
    \caption{An illustration of how the typical estimate of the $\mmd$ is unreliable especially as $\e$ increases. In addition, this result matches our intuition from the Bootstrap method; as $\e$ increases, the two groups become increasingly mixed and more similar, and the $\mmd$ approaches $0$.} \label{fig:typical_est}
\end{figure}

Figure~\ref{fig:typical_est} illustrates the that the typical estimate of the $\mmd$ (equation~\ref{eq:typical_est}) is unreliable, especially as $\e$ increases. It also demonstrates the upper and lower bounds of S-SD as simulated epsilon contamination increases; S-SD bounds contain the true value of the $\mmd$ at all values of $\e$.

Table \ref{tab:n2000} shows the same results as those presented in table~\ref{nonrandom_main_table} with $N=2000$ instead of $N=100$. The results show that, as seen in figure ~\ref{IW_FCR}, the performance of QNO and S-QNO improves as sample size increases, while S-SD continues to have tight and informative bounds.

\begin{table}[H]
\caption{MIW and FCR for MIMIC and FOREST at $\e = 0.2$. }
\resizebox{\textwidth}{!}{
\begin{tabular}{l|ll|ll|}
\toprule
          & \multicolumn{2}{c|}{MIMIC ($n = 2000, d = 2)$} & \multicolumn{2}{c|}{FOREST $(n = 2000, d = 54)$} \\
\cmidrule(r){2-3} \cmidrule(r){4-5}
Approach  & FCR    & MIW   & FCR    & MIW    \\
\cmidrule(r){2-3} \cmidrule(r){4-5}
S-SD (Ours)       & $0.0 \pm (0.0)$ & $0.251 \pm (0.008)$   & $0.0 \pm (0.0)$ & $0.128 \pm (0.007)$ \\
S-QNO     & $0.0 \pm (0.0)$ & $0.25 \pm (0.006)$     & $0.0 \pm (0.0)$ & $0.134 \pm (0.007)$  \\
QNO               & $0.0 \pm (0.0)$ & $0.227 \pm (0.006)$   & $0.32 \pm (0.066)$ & $0.107 \pm (0.01)$\\
SD               & $0.02 \pm (0.02)$ & $0.23 \pm (0.009)$   & $0.46 \pm (0.07)$ & $0.087 \pm (0.009)$   \\
SM               & $0.02 \pm (0.02)$ & $0.217 \pm (0.008)$  & $0.46 \pm (0.07)$ & $0.081 \pm (0.008)$ \\
Bootstrap        & $0.3 \pm (0.065)$ & $0.091 \pm (0.004)$  & $0.46 \pm (0.07)$ & $0.042 \pm (0.003)$  \\
\bottomrule
\label{tab:n2000}
\end{tabular}}
\end{table}

\subsection{Additional results using MIMIC data}
\begin{figure}[H]
    \centering
    \includegraphics[width=\textwidth]{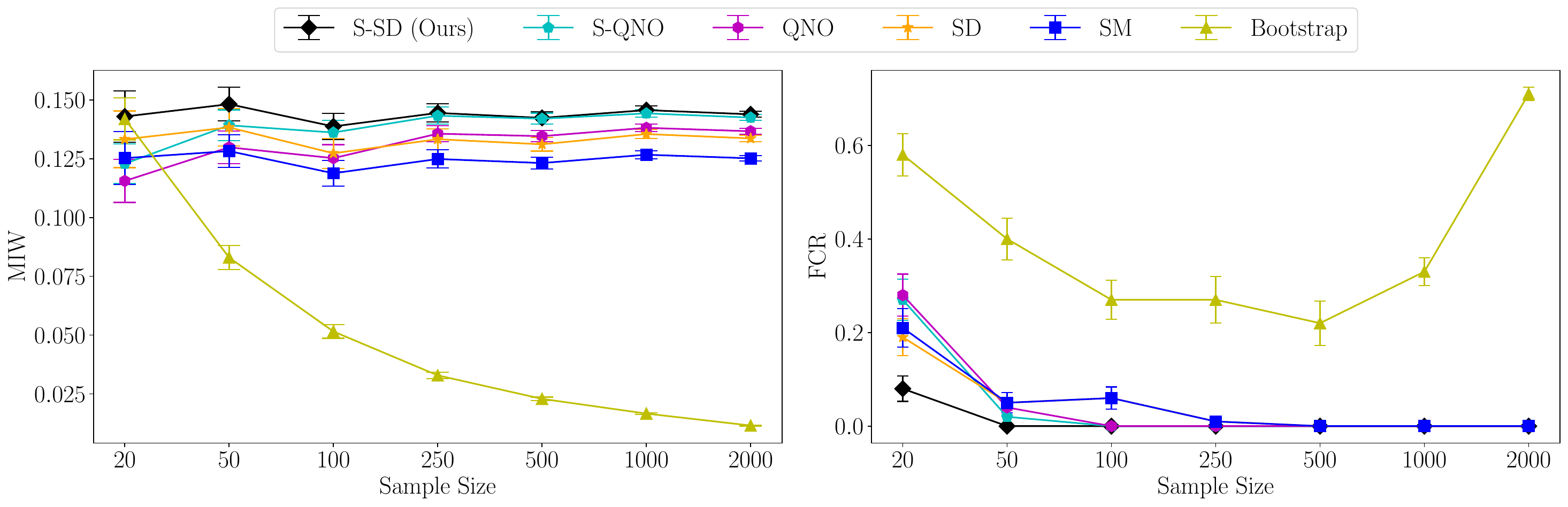}
    \caption{The same experiment as in figure \ref{IW_FCR}, but run in the MIMIC $(n=100,d=2)$ setting.}\label{fig:n_mimic}
\end{figure}

\begin{figure}[H]
    \centering
    \includegraphics[width=\textwidth]{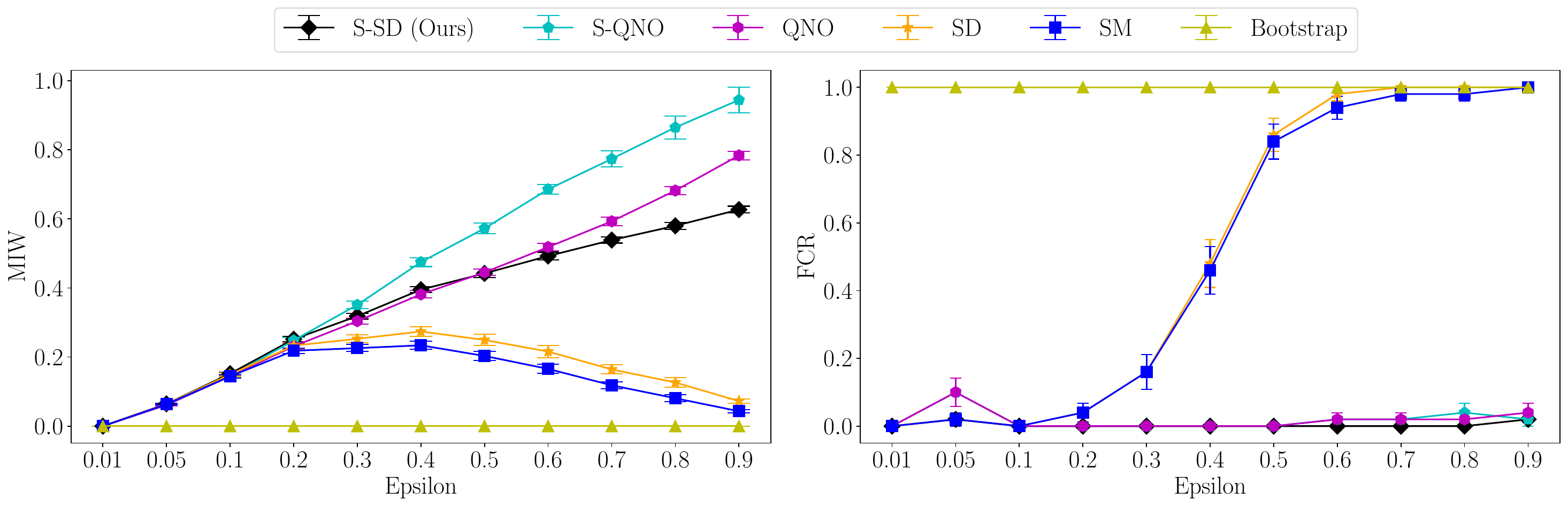}
    \caption{The same experiment as in figure \ref{fig:FOREST_eps}, but run in the MIMIC $(n=100,d=2)$ setting.}\label{fig:eps_mimic}
\end{figure}

Figures \ref{fig:n_mimic} and \ref{fig:eps_mimic} are similar to figures~\ref{IW_FCR} and ~\ref{fig:FOREST_eps} in the main text, but instead of performing the analysis on the FOREST data, we perform the analysis on the MIMIC data. The results are largely consistent with the analysis in the main text: our approach outperforms others in that it gives the lowest FCR for every sample size and every value of $\e$.

\subsection{Additional results using BIO data}
\begin{figure}[H]
    \centering
    \includegraphics[width=\textwidth]{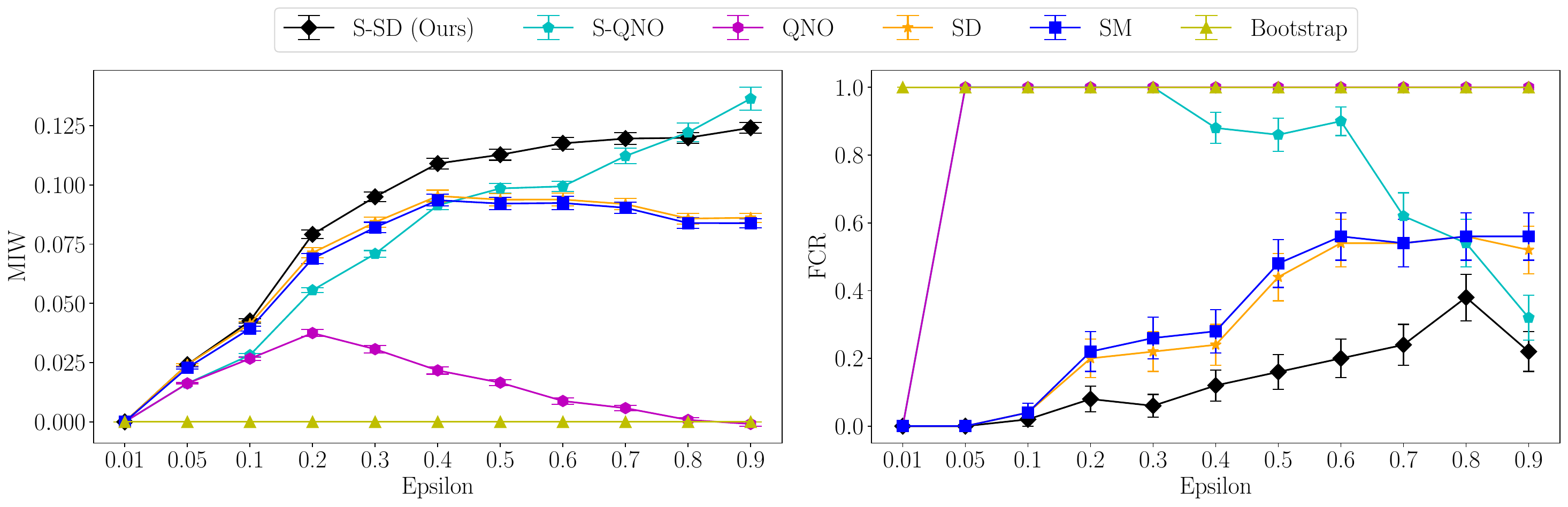}
    \caption{The same experiment as in figure \ref{fig:FOREST_eps}, but run in the BIO $(n=72,d=7128)$ setting.}\label{fig:eps_bio}
\end{figure}

Figure \ref{fig:eps_bio} is similar to figure \ref{fig:FOREST_eps} in the main text, but instead of performing the analysis on the FOREST data, we perform the analysis on the BIO data. We note that due to the limited sample size of the BIO data, we are unable to create figure~\ref{IW_FCR} for the BIO data. S-SD gives the lowest FCR for every value of $\e$. As in \ref{nonrandom_main_table}, QNO and S-QNO have a irreducible dependence on the dimension size of the data. QNO fails to contain the value of the true $\mmd$ at all $\e \ge 0.01$. S-QNO performs poorly until larger values of epsilon, where the step approximation becomes effective; this is because at small sample sizes, the set of corrupted samples is small, and the approximation cannot be divided into many steps.

\subsection{Step Size Sensitivity}
Table ~\ref{tab:sens_both} shows that similar to S-SD, S-QNO gives bound estimates with FCR of zero even for a few number of steps. Conclusions from the main text regarding setting the step size for S-SD hold for S-QNO as well.

\begin{table}[H] \label{MIMIC_FOREST_2k}
\centering
\begin{tabular}{c|ll|}
\toprule
 & \multicolumn{2}{c|}{S-QNO} \\
\cmidrule(r){2-3} 
Number of Steps  & FCR    & MIW  \\
\cmidrule(r){2-3} 
$2$  & $0.05 \pm (0.023$) & $0.067 \pm (0.001)$ \\
$3$ & $0.0 \pm (0.0)$ & $0.066 \pm (0.001)$ \\
$5$  & $0.0 \pm (0.0)$ & $0.075 \pm (0.0)$ \\
$10$  & $0.0 \pm (0.0)$ & $0.091 \pm (0.001)$ \\
$20$  & $0.0 \pm (0.0)$ & $0.087 \pm (0.001)$ \\
$50$  & $0.0 \pm (0.0)$ & $0.091 \pm (0.001)$ \\
\bottomrule
\end{tabular}
\caption{Varying number of steps for S-QNO in FOREST $(n = 2000, d = 54)$ with $\e = 0.2$. Standard errors (shown in parentheses) represent the SE for the FCR and MIW for each method over 100 trials. In each trial, we sample 2000 data points without replacement and simulate $\e$-contamination, and then compute the bounds for S-QNO at each number of steps on the same sample.}\label{tab:sens_both}
\end{table}

\section{Experimental results from the random contamination setting}

\begin{figure}[H]
    \centering
    \includegraphics[width=\textwidth]{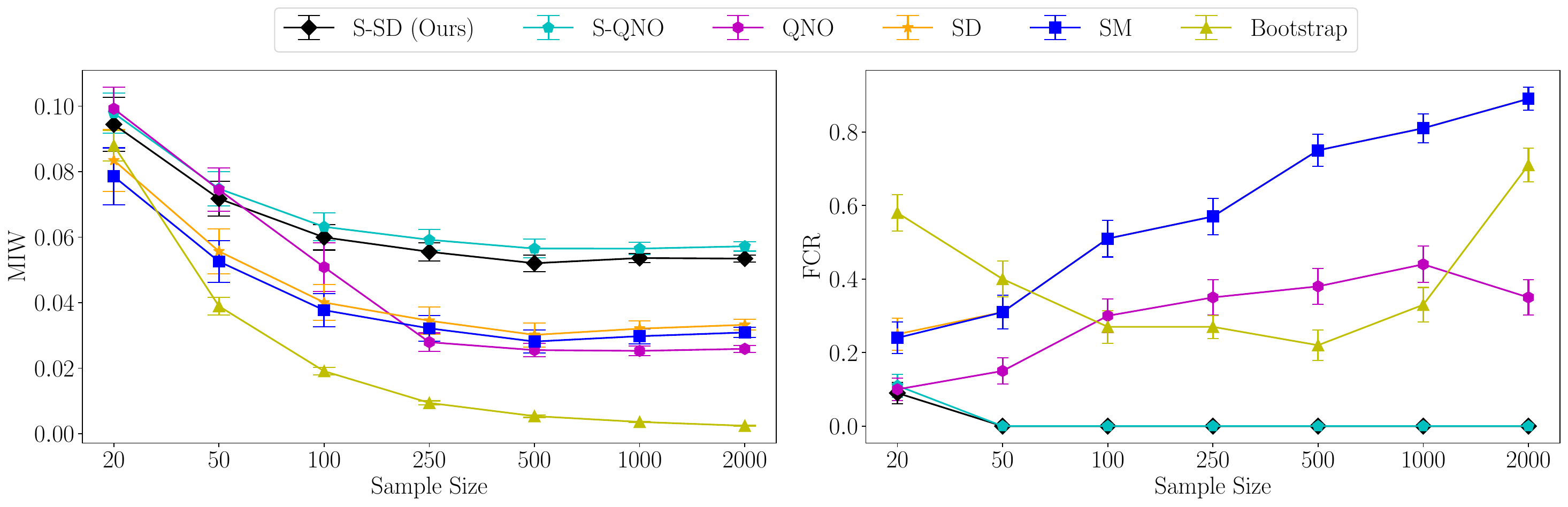}
    \caption{The MIW and FCR for each approach is shown as the sample size increases when $\e = 0.2$ in FOREST $(n=100, d=54)$. Bars indicate the SE of the FCR and MIW across all trials.}
    \label{fig:randsamp}
\end{figure}

\begin{figure}[H]
    \centering
    \includegraphics[width=\textwidth]{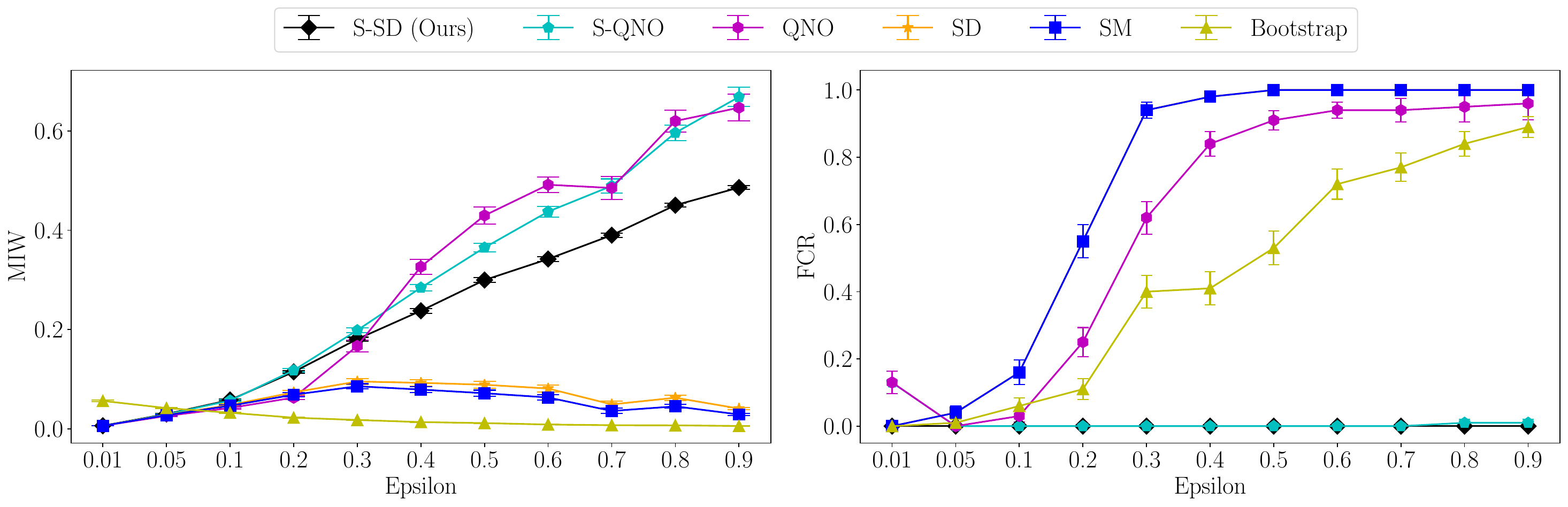}
    \caption{The MIW and FCR for each approach is shown as the intensity of random $\e$-contamination varies from $\e = 0.01$ to $\e = 0.9$ in FOREST $(n=100, d=54)$. Bars indicate the SE of the FCR and MIW across all trials.}
    \label{fig:rand_eps}
\end{figure}

We present the same experiments as in table ~\ref{nonrandom_main_table} and figures ~\ref{IW_FCR} and ~\ref{fig:FOREST_eps} on FOREST  $(n = 100, d = 54)$ when the set of contaminations $C^*$ is a random sample of $X$ of size $\lfloor \e n \rfloor$, rather than the $\lfloor \e n \rfloor$ samples in $X$ with the largest witness function values as described in section \ref{Setup}. The results in table ~\ref{tab:rand100} and figure ~\ref{fig:randsamp} are consistent with the results in the main text and show that for all $\e$, S-SD gives the most credible bounds with the tightest MIW. Figure ~\ref{fig:rand_eps} shows that FCR and MIW decrease for S-SD, S-QNO, and QNO as sample size increases in FOREST.

\begin{table}[H]
\resizebox{\textwidth}{!}{
\begin{tabular}{l|ll|ll|ll|}
\toprule
          & \multicolumn{2}{c|}{MIMIC ($n = 100, d = 2)$} & \multicolumn{2}{c|}{FOREST $(n = 100, d = 54)$} & \multicolumn{2}{c|}{BIO $(n = 72, d = 7128)$} \\
\cmidrule(r){2-3} \cmidrule(r){4-5} \cmidrule(r){6-7}
Approach  & FCR    & MIW   & FCR    & MIW & FCR    & MIW    \\
\cmidrule(r){2-3} \cmidrule(r){4-5} \cmidrule(r){6-7}
S-SD (Ours)    & $0.0 \pm (0.0)$ & $0.258 \pm (0.002)$  & $0.0 \pm (0.0)$ & $0.107 \pm (0.002)$  & $0.07 \pm (0.026)$ & $0.08 \pm (0.001)$ \\
S-QNO    & $0.0 \pm (0.0)$ & $0.258 \pm (0.002)$  & $0.0 \pm (0.0)$ & $0.114 \pm (0.002)$   & $1.0 \pm (0.0)$ & $0.056 \pm (0.001)$ \\
QNO               & $0.0 \pm (0.0)$ & $0.247 \pm (0.002)$  & $0.4 \pm (0.069)$ & $0.051 \pm (0.002)$  & $1.0 \pm (0.0)$ & $0.038 \pm (0.001)$ \\
SD                & $0.0 \pm (0.0)$ & $0.24 \pm (0.002)$  & $0.92 \pm (0.038)$ & $0.064 \pm (0.003)$  & $0.15 \pm (0.036)$ & $0.074 \pm (0.001)$ \\
SM               & $0.0 \pm (0.0)$ & $0.225 \pm (0.002)$  & $0.92 \pm (0.038)$ & $0.06 \pm (0.003)$   & $0.42 \pm (0.049)$ & $0.05 \pm (0.002)$ \\
Bootstrap       & $1.0 \pm (0.0)$ & $0.02 \pm (0.0)$    & $0.6 \pm (0.069)$ & $0.005 \pm (0.0)$  & $0.85 \pm (0.036)$ & $0.036 \pm (0.001)$ \\
\bottomrule
\label{tab:rand100}
\end{tabular}}
\caption{100 Samples random contaminations}
\end{table}

\end{document}